\documentclass{article}
\usepackage{microtype}
\usepackage[pdftex]{graphicx}
\usepackage{subfigure}
\usepackage{booktabs} 
\usepackage[frozencache=true,cachedir=minted-cache]{minted}
\usepackage{amssymb}
\usepackage{pifont}
\usepackage{wrapfig}
\usepackage{nonfloat}
\usepackage{caption}[=v1]
\usepackage{url}

\usepackage{hyperref}
\usepackage{xspace}

\usepackage[accepted]{icml2024}

\usepackage{amsmath}
\usepackage{amssymb}
\usepackage{mathtools}
\usepackage{amsthm}
\usepackage{booktabs}
\usepackage{multirow}
\usepackage{listings}
\usepackage{xcolor}
\usepackage{tcolorbox}
\tcbuselibrary{listings}
\usepackage{tcolorbox}
\tcbuselibrary{listings}

\definecolor{lightgray}{gray}{0.2}  

\usepackage[capitalize,noabbrev]{cleveref}

\theoremstyle{plain}

\theoremstyle{definition}

\theoremstyle{remark}

\usepackage[textsize=tiny]{todonotes}

\icmltitlerunning{A Multimodal Automated Interpretability Agent}
\newcommand{\milan}{\textsc{milan}\xspace}
\newcommand{\milannotations}{\textsc{milannotations}\xspace}
\newcommand{\maia}{\textsc{maia}\xspace}
\setmintedinline{fontsize=\small, breaklines=true}

\newcommand{\eg}{\emph{e.g.}\xspace}
\newcommand{\ie}[1]{\emph{i.e}}

\definecolor{boxgray}{rgb}{.94, .94, .94}

\newcommand{\code}[1]{\mintinline{python}|#1|}
\newcommand{\myparagraph}[1]{\vspace{0pt} \noindent \textbf{#1}}

\lstdefinestyle{mystyle}{
  basicstyle=\ttfamily\scriptsize,
  breaklines=true,
  columns=fullflexible,
  breakindent=0pt,
}

\newtcblisting{mygraybox}[2][]{
  arc=1mm,
  colback=boxgray,  
  listing only,
  listing style=mystyle,
  title=#2,
  boxrule=.08mm,  
  boxsep=.1mm,
  #1
}

\lstdefinestyle{mystylepython}{
    language=Python,
    basicstyle=\ttfamily\scriptsize,
    otherkeywords={self},             
    keywordstyle=\ttfamily\scriptsize\color{blue},
    emph={MyClass,__init__},          
    emphstyle=\ttfamily\scriptsize\color{blue},    
    stringstyle=\color{blue}, 
    commentstyle=\color{gray},   
    showstringspaces=false,            
    breaklines = true
    backgroundcolor=\color{lightgray}, 
}

\begin{document}

\twocolumn[
\icmltitle{A Multimodal Automated Interpretability Agent}



\icmlsetsymbol{equal}{*}

\begin{icmlauthorlist}
\icmlauthor{Tamar Rott Shaham}{csail,equal}
\icmlauthor{Sarah Schwettmann}{csail,equal}\\
\icmlauthor{Franklin Wang}{csail}
\icmlauthor{Achyuta Rajaram}{csail}
\icmlauthor{Evan Hernandez}{csail}
\icmlauthor{Jacob Andreas}{csail}
\icmlauthor{Antonio Torralba}{csail}
\end{icmlauthorlist}

\icmlaffiliation{csail}{MIT CSAIL}

\icmlcorrespondingauthor{Tamar Rott Shaham}{tamarott@mit.edu}
\icmlcorrespondingauthor{Sarah Schwettmann}{schwett@mit.edu}


\vskip 0.3in
]





\begin{abstract}

This paper describes \maia, a \textbf{M}ultimodal \textbf{A}utomated \textbf{I}nterpretability \textbf{A}gent. \maia is a system that uses neural models to automate neural model understanding tasks like feature interpretation and failure mode discovery. It equips a pre-trained vision-language model with a set of tools that support iterative experimentation on subcomponents of other models to explain their behavior. These include tools commonly used by human interpretability researchers: for synthesizing and editing inputs, computing maximally activating exemplars from real-world datasets, and summarizing and describing experimental results. 
\textit{Interpretability experiments} proposed by \maia compose these tools to describe and explain system behavior. 
We evaluate applications of \maia to computer vision models.
We first characterize \maia's ability to describe (neuron-level) features in learned representations of images. Across several trained models and a novel dataset of synthetic vision neurons with paired ground-truth descriptions, \maia produces descriptions comparable to those generated by expert human experimenters.
We then show that \maia can aid in two additional interpretability tasks: reducing sensitivity to spurious features, and automatically identifying inputs likely to be mis-classified.\textsuperscript{\href{\#footnote}{\dag}}
\end{abstract}
\renewcommand{\thefootnote}{\fnsymbol{footnote}}


\setcounter{footnote}{1}

\renewcommand{\thefootnote}{\dag}
\footnotetext{\label{footnote}Website: \url{https://multimodal-interpretability.csail.mit.edu/maia}}
\vspace{-.5cm}
\section{Introduction}

Understanding of a neural model can take many forms. Given an image classifier, for example, we may wish to recognize when and how it relies on sensitive features like race or gender, identify systematic errors in its predictions, or learn how to modify the training data and model architecture to improve accuracy and robustness. Today, this kind of understanding requires significant effort on the part of researchers---involving exploratory data analysis, formulation of hypotheses, and controlled experimentation \citep{nushi2018towards, zhang2018manifold}. As a consequence, this kind of understanding is slow and expensive to obtain even about the most widely used models.

\begin{figure}[t!]
    \centering \includegraphics[width=.93\linewidth]{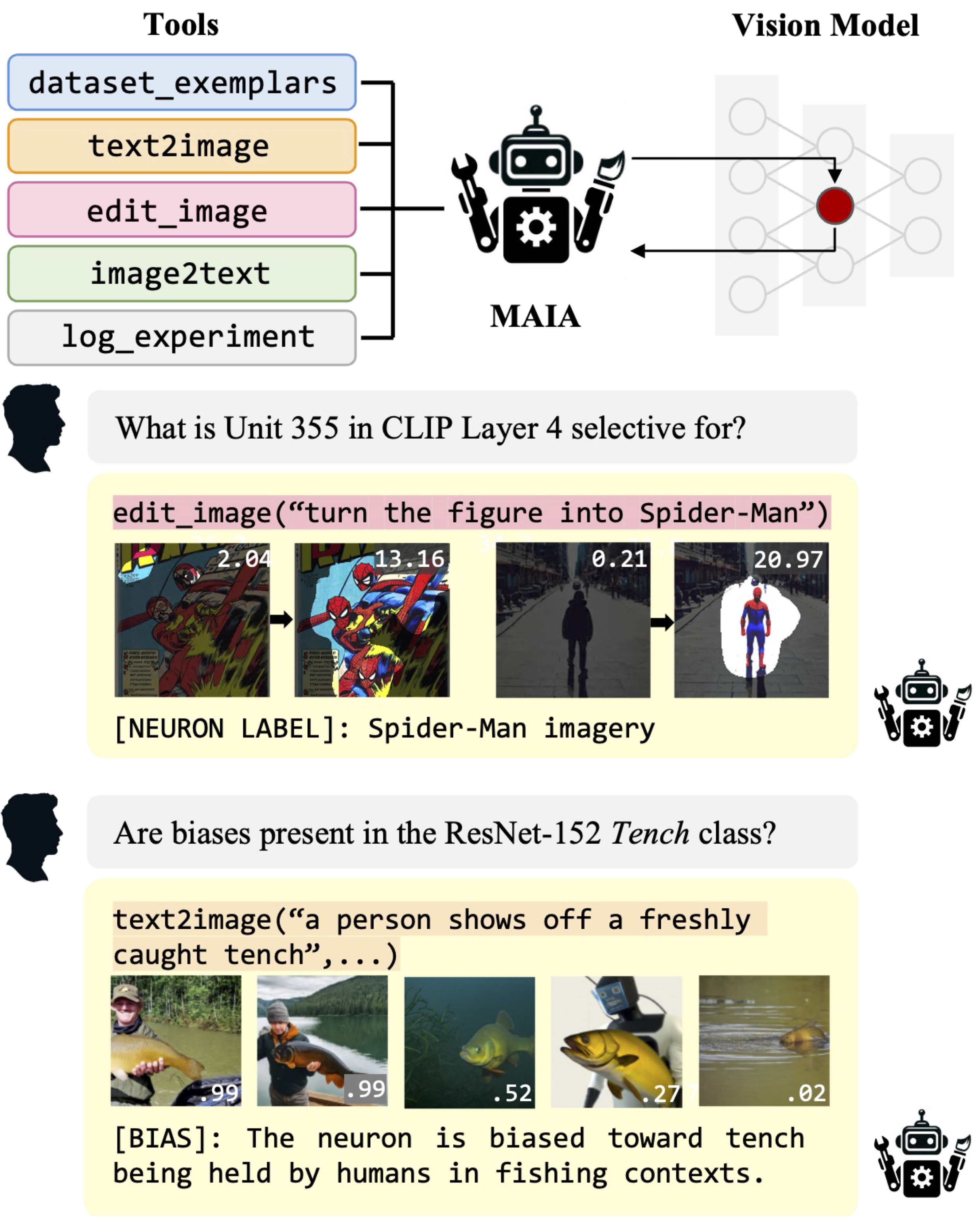}
    \vspace{-2mm}
    \caption{\textbf{\maia framework.} \maia autonomously conducts experiments on other systems to explain their behavior.}
    \vspace{-6mm}
    \label{fig:teaser}
\end{figure}

\begin{figure*}[htbp]
    \centering
    \includegraphics[width=\textwidth]{figures/MAIA_main_CR.pdf}
    \caption{\textbf{\maia experiments for labeling neurons.} \maia iteratively writes programs that compose common interpretability tools to conduct experiments on other systems. At each step, \maia autonomously makes and updates hypotheses in light of experimental outcomes, showing sophisticated scientific reasoning capabilities. Generated code is executed with a Python interpreter and the outputs (shown above, neuron activation values overlaid in white, masks thresholded at 0.95 percentile of activation maps) are returned to \maia. 
    }
    \label{fig:examples}
\end{figure*}

Recent work on \textbf{automated interpretability} \citep[\eg][]{hernandez2022natural,bills2023language,schwettmann2023find} has begun to address some of these limitations by using \emph{learned models themselves} to assist with model understanding tasks---for example, by assigning natural language descriptions to learned representations, which may then be used to surface features of concern. But current methods are useful almost exclusively as tools for hypothesis generation; they characterize model behavior on a limited set of inputs, and are often low-precision \citep{huang2023rigorously}. How can we build tools that help users understand models, while combining the flexibility of human experimentation with the scalability of automated techniques?


This paper introduces a prototype system we call the \textbf{M}ultimodal \textbf{A}utomated \textbf{I}nterpretability \textbf{A}gent (\maia), which combines a pretrained vision-language model backbone with an API containing tools designed for conducting experiments on deep networks. \maia is prompted with an explanation task (\eg ``describe the behavior of unit 487 in layer 4 of CLIP'' or ``in which contexts does the model fail to classify \code{labradors}?'') and designs an \textit{interpretability experiment} that composes experimental modules to answer the query. \maia's modular design (Figure \ref{fig:teaser}) enables flexible evaluation of arbitrary systems and straightforward incorporation of new experimental tools. Section \ref{sec:framework} describes the current tools in \maia's API, including modules for synthesizing and editing novel test images, which enable direct hypothesis testing during the interpretation process. 

We evaluate \maia's ability to produce predictive explanations of vision system components using the neuron description paradigm \citep{netdissect2017, bau2020units, oikarinen2022clip, bills2023language, singh2023explaining, schwettmann2023find} which appears as a subroutine of many interpretability procedures. We additionally introduce a novel dataset of \textit{synthetic vision neurons} built from an open-set concept detector with ground-truth selectivity specified via text guidance. In Section \ref{sec:evaluation}, we show that \maia desriptions of both synthetic neurons and neurons in the wild are more predictive of behavior than baseline description methods, and in many cases on par with human labels. 

\maia also automates model-level interpretation tasks where descriptions of learned representations produce actionable insights about model behavior. We show in a series of experiments that \maia's iterative experimental approach can be applied to downstream model auditing and editing tasks including spurious feature removal and bias identification in a trained classifier. Both applications demonstrate the adaptability of the \maia framework across experimental settings: novel end-use cases are described in the user prompt to the agent, which can then use its API to compose programs that conduct task-specific experiments. While these applications show preliminary evidence that procedures like \maia which automate both experimentation and description have high potential utility to the interpretability workflow, we find that \maia still requires human steering to avoid common pitfalls including confirmation bias and drawing conclusions from small sample sizes. Fully automating end-to-end interpretation of other systems will not only require more advanced tools, but agents with more advanced capabilities to reason about how to use them.
\section{Related work}

\paragraph{Interpreting deep features.}  
Investigating individual neurons inside deep networks reveals a range of human-interpretable features. Approaches to describing these neurons use exemplars of their behavior as explanation, either by visualizing features they select for \citep{zeiler2014visualizing, girshick2014rich, karpathy2015visualizing,mahendran2015understanding,olah2017feature} or automatically categorizing maximally-activating inputs from real-world datasets \citep{netdissect2017,bau2020units, oikarinen2022clip, dalvi2019one}. Early approaches to translating visual exemplars into language descriptions drew labels from fixed vocabularies \cite{netdissect2017}, or produced descriptions in the form of programs \citep{mu2021compositional}.
\vspace{-1mm}
\paragraph{Automated interpretability.}
Later work on automated interpretability produced open-ended 
descriptions of learned features in the form of natural language text, either curated from human labelers \citep{schwettmann2021toward} or generated directly by learned models \citep{hernandez2022natural, bills2023language, gandelsman2024interpreting}. However, these labels are often unreliable as \emph{causal} descriptions of model behavior without further experimentation \citep{huang2023rigorously}. \citet{schwettmann2023find} introduced the Automated Interpretability Agent protocol for experimentation on black-box systems using a language model agent, though this agent operated purely on language-based exploration of inputs, which limited its action space. \maia similarly performs iterative experimentation rather than labeling features in a single pass, but has access to a library of interpretability tools as well as built-in vision capabilities. \maia's modular design also supports experiments at different levels of granularity, ranging from analysis of individual features to sweeps over entire networks, or identification of more complex network subcomponents \citep{conmy2023towards}.

\vspace{-1mm}
\paragraph{Language model agents.}

Modern language models are promising foundation models for interpreting other networks due to their strong reasoning capabilities \citep{openai2023gpt4}. These capabilities can be expanded by using the LM as an \emph{agent}, where it is prompted with a high-level goal and has the ability to call external tools such as calculators, search engines, or other models in order to achieve it \cite{schick2023toolformer, qin2023toolllm}. When additionally prompted to perform chain-of-thought style reasoning between actions, agentic LMs excel at multi-step reasoning tasks in complex environments \cite{yao2023react}.
\maia leverages an agent architecture to generate and test hypotheses about neural networks trained on vision tasks. While ordinary LM agents are generally restricted to tools with textual interfaces, previous work has supported interfacing with the images through code generation \citep{surís2023vipergpt, wu2023visual}. More recently, large multimodal LMs like GPT-4V have enabled the use of image-based tools directly \citep{zheng2024gpt4vision,chen2023endtoend}. \maia follows this design and is, to our knowledge, the first multimodal agent equipped with tools for interpreting deep networks.
\section{\maia Framework}
\label{sec:framework}

\maia is an agent that autonomously conducts experiments on other systems to explain their behavior, by composing interpretability subroutines into Python programs. Motivated by the promise of using language-only models to complete one-shot visual reasoning tasks by calling external tools \citep{surís2023vipergpt, gupta2023visual}, and the need to perform \textit{iterative} experiments with both visual and numeric results, we build \maia from a pretrained multimodal model with the ability to process images directly. \maia is implemented with a GPT-4V vision-language model (VLM) backbone \citep{openai2023gpt4v} . Given an interpretability query (\eg \textit{Which neurons in Layer 4 are selective for forested backgrounds?}), \maia runs experiments that test specific hypotheses (\eg computing neuron outputs on images with edited backgrounds), observes experimental outcomes, and updates hypotheses until it can answer the user query. 

We enable the VLM to design and run interpretability experiments using the \maia API, which defines two classes: the \code{System} class and the \code{Tools} class, described below. The API is provided to the VLM in its system prompt. We include a complete API specification in Appendix \ref{sec:API_appendix}. The full input to the VLM is the API specification followed by a ``user prompt'' describing a particular interpretability task, such as explaining the behavior of an individual neuron inside a vision model with natural language (see Section \ref{sec:evaluation}). To complete the task, \maia uses components of its API to write a series of Python programs that run experiments on the system it is interpreting. \maia outputs function definitions as strings, which we then execute internally using the Python interpreter. The Pythonic implementation enables flexible incorporation of built-in functions and existing packages, \eg the \maia API uses the PyTorch library \citep{paszke2019pytorch} to load common pretrained vision models.
\vspace{-2mm}
\subsection{System API}
The \code{System} class inside the \maia API instruments the system to be interpreted and makes subcomponents of that system individually callable. For example, to probe single neurons inside ResNet-152 \citep{he2016deep}, \maia can use the \code{System} class to initialize a neuron object by specifying its number and layer location, and the model that the neuron belongs to: \code{system = System(unit_id, layer_id, model_name)}. \maia can then design experiments that test the neuron's activation value on different image inputs by running \code{system.neuron(image_list)}, to return activation values and masked versions of the images in the list that highlight maximally activating regions (See Figure \ref{fig:examples} for examples). While existing approaches to common interpretability tasks such as neuron labeling require training specialized models on task-specific datasets \citep{hernandez2022natural}, the \maia system class supports querying arbitrary vision systems without retraining.

\subsection{Tool API}

The \code{Tools} class consists of a suite of functions enabling \maia to write modular programs that test hypotheses about system behavior. \maia tools are built from common interpretability procedures such as characterizing neuron behavior using real-world images \citep{netdissect2017} and performing causal interventions on image inputs \citep{hernandez2022natural, casper2022diagnostics}, which \maia then composes into more complex experiments (see Figure \ref{fig:examples}). When programs written by \maia are compiled internally as Python code, these functions can leverage calls to other pretrained models to compute outputs. For example, \code{tools.text2image(prompt_list)} returns synthetic images generated by a text-guided diffusion model, using prompts written by \maia to test a neuron's response to specific visual concepts.
The modular design of the tool library enables straightforward incorporation of new tools as interpretability methods grow in sophistication. For the experiments in this paper we use the following set:

\myparagraph{Dataset exemplar generation.}
Previous studies have shown that it is possible to characterize the prototypical behavior of a neuron by recording its activation values over a large dataset of images~\cite{netdissect2017,bau2020units}. We give \maia the ability to run such an experiment on the validation set of ImageNet~\cite{deng2009imagenet} and construct the set of 15 images that maximally activate the system it is interpreting. Interestingly, \maia often chooses to begin experiments by calling this tool (Figure \ref{fig:examples}). We analyze the importance of the \code{dataset_exemplars} tool in our ablation study (\ref{sec:ablation}).  

\myparagraph{Image generation and editing tools.} \maia is equipped with a \code{text2image(prompts)} tool that synthesizes images by calling Stable Diffusion v1.5~\cite{Rombach_2022_CVPR} on text prompts. Generating inputs enables \maia to test system sensitivity to fine-grained differences in visual concepts, or test selectivity for the same visual concept across contexts (\eg the \textit{bowtie} on a pet and on a child in Figure \ref{fig:examples}). We analyze the effect of using different text-to-image models in Section~\ref{sec:ablation}. In addition to synthesizing new images, \maia can also edit images using Instruct-Pix2Pix~\cite{brooks2022instructpix2pix} by calling \code{edit_images(image, edit_instructions)}. Generating and editing synthetic images enables hypothesis tests involving images lying outside real-world data distributions, \eg the addition of antelope horns to a horse (Figure \ref{fig:examples}, see \textit{Causal intervention on image input}). 

\begin{figure*}[t]
    \centering
    \includegraphics[width=0.9\linewidth]{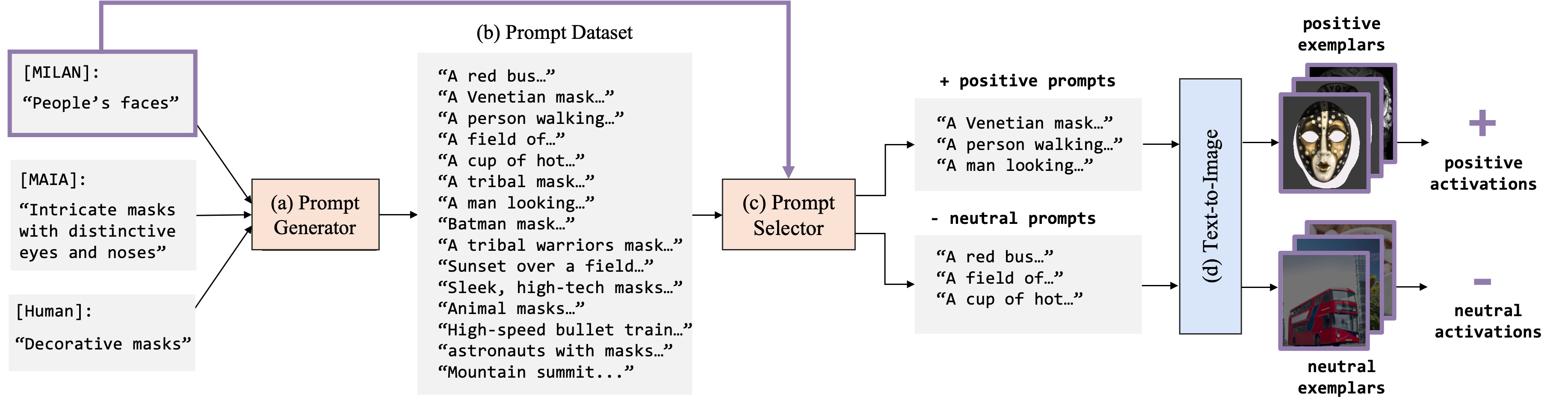}
    \vspace{-3mm}
    \caption{\textbf{Predictive evaluation protocol}. 
    We compare neuron labeling methods by assessing how well their labels predict neuron activation values on unseen data. For each neuron we perform the following steps:
    (a) An LM uses candidate neuron labels to generate a set of image prompts that should maximally/neutrally activate the neuron. 
    (b) All prompts (positive and neutral) from all methods are combined into one dataset. (c) For each labeling method, a new LM selects prompts from the Prompt Dataset that are likely to produce maximal and neutral neuron activations, if that label were accurate. (d) A text-to-image model generates all corresponding images, and the average activation values for positive and neutral images are recorded. A predictive neuron label will produce exemplars with maximally positive activations relative to the neutral baseline.
    }
    \vspace{-4mm}
    \label{fig:eval_schematic}
\end{figure*}

\vspace{-2mm}
\paragraph{Image description and summarization tools.} To limit confirmation bias in \maia's interpretation of experimental results, we use a multi-agent framework in which \maia can ask a new instance of GPT-4V with no knowledge of experimental history to describe highlighted image regions in individual images, \code{describe_images(image_list)}, or summarize what they have in common across a group of images, \code{summarize_images(image_list)}. We observe that \maia uses this tool in situations where previous hypotheses failed or when observing complex combinations of visual content. 

\vspace{-2mm}
\paragraph{Experiment log.}
\maia can document the results of each experiment (\eg images, activations) using the \code{log_experiment} tool, to make them accessible during subsequent experiments. We prompt \maia to finish experiments by logging results, and let it choose what to log (\eg data that clearly supports or refutes a particular hypothesis).

\section{Evaluation}
\label{sec:evaluation}
 
The \maia framework is task-agnostic and can be adapted to new applications by specifying an interpretability task in the user prompt to the VLM. Before tackling model-level interpretability problems (Section \ref{sec:experiments}), we evaluate \maia's performance on the black-box neuron description task, a widely studied interpretability subroutine that serves a variety of downstream model auditing and editing applications \citep{gandelsman2024interpreting, yang2023language, hernandez2022natural}. For these experiments, the user prompt to \maia specifies the task and output format (a longer-form \code{[DESCRIPTION]} of neuron behavior, followed by a short \code{[LABEL]}), and \maia's \code{System} class instruments a particular vision model (\eg \code{ResNet-152}) and an individual unit indexed inside that model (\eg \code{Layer 4 Unit 122}). Task specifications for these experiments may be found in Appendix \ref{sec:user_appendix}. We find \maia correctly predicts behaviors of individual vision neurons in three trained architectures (Section \ref{sec:real_eval}), and in a synthetic setting where ground-truth neuron selectivities are known (Section \ref{sec:syn_eval}). We also find descriptions produced by \maia's interactive procedure to be more predictive of neuron behavior than descriptions of a fixed set of dataset exemplars, using the \milan baseline from \citet{hernandez2022natural}. In many cases, \maia descriptions are on par with those by human experts using the \maia API. In Section \ref{sec:ablation}, we perform ablation studies to test how components of the \maia API differentially affect description quality.

\subsection{Neurons in vision models}
\label{sec:real_eval}

We use \maia to produce natural language descriptions of a subset of neurons across three vision architectures trained under different objectives: ResNet-152, a CNN for supervised image classification \citep{he2016deep}, DINO \citep{caron2021emerging}, a Vision Transformer trained for unsupervised representation learning \citep{grill2020bootstrap, chen2021exploring}, and the CLIP visual encoder \citep{radford2021learning}, a ResNet-50 model trained to align image-text pairs. For each model, we evaluate descriptions of $100$ units randomly sampled from a range of layers that capture features at different levels of granularity (ResNet-152 \textit{conv.}1, \textit{res.}1-4, DINO MLP 1-11, CLIP \textit{res.}1-4). Figure \ref{fig:examples} shows examples of \maia experiments on neurons from all three models, and final \maia labels. We also evaluate a baseline non-interactive approach that only labels dataset exemplars of each neuron's behavior using the \milan model from \citet{hernandez2022natural}. Finally, we collect human annotations of a random subset (25\%) of neurons labeled by \maia and \milan, in an experimental setting where human experts write programs to perform interactive analyses of neurons using the \maia API. Human experts receive the \maia user prompt, write programs that run experiments on the neurons, and return neuron descriptions in the same format. See Appendix \ref{sec:human_eval} for details on the human labeling experiments. 

\begin{figure}[t]
    \centering
    \includegraphics[width=1\columnwidth]{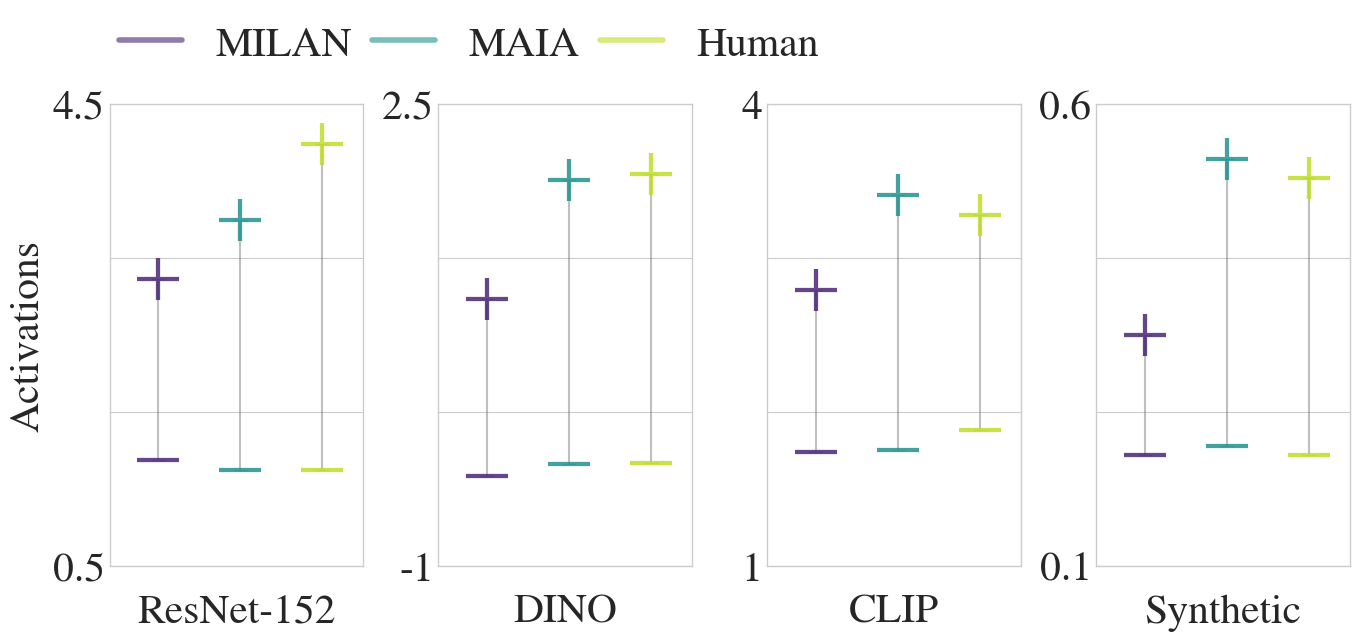}
    \vspace{-6mm}
    \caption{\textbf{Predictive evaluation results.} The average positive activation values (``+'') for \maia labels outperform \milan and are comparable to human descriptions for both real and synthetic neurons. Neutral activations (``-'') are comparable across methods.}
    \label{fig:eval-results}
    \vspace{-4mm}
\end{figure}

We evaluate the accuracy of neuron descriptions produced by \maia, \milan, and human experts by measuring how well they \textit{predict} neuron behavior on unseen test images (Figure \ref{fig:eval_schematic}). Similar to evaluation approaches that produce contrastive or counterfactual exemplars to reveal model decision boundaries \citep{gardner2020evaluating, kaushik2020learning}, we use candidate neuron labels to generate new images that should elicit maximally positive activations relative to a neutral baseline. For a given neuron, we generate a pool of image candidates by providing \maia, \milan, and human labels to a Prompt Generator model (implemented with a new instance of GPT-4). For each candidate label (\eg \textit{intricate masks}), the Prompt Generator is instructed to write 7 image prompts that should generate maximally activating images (\eg \textit{A Venetian mask, A tribal mask,...}), and 7 prompts for neutral images (unrelated to the label) that should elicit baseline activations (\eg \textit{A red bus, A field of flowers,...}). All positive and neutral prompts from all labeling methods (\maia, \milan, and human experts) form a Prompt Dataset of 42 prompts per neuron. Next, we evaluate the accuracy of each candidate label by using a Prompt Selector LM (implemented with another GPT-4 instance) to match that label  with the 7 prompts it is most and least likely to entail. We then generate the corresponding images using a text-to-image model (DALL-E3) and measure neuron activation values on those images. If a neuron label is predictive of activations, it will be matched with positive exemplars that maximally activate the neuron relative to the neutral baseline. Combining prompts from all methods into one test set (vs. evaluating each model separately) more rigorously evaluates the completeness of each candidate label: an incomplete description produced by one labeling method (e.g. \textit{trains} for a neuron selective for \textit{trains} \verb|OR| \textit{dogs}) could be matched with a ``neutral'' image prompt describing \textit{dogs}, which would in fact elicit high activation. This method primarily discriminates between labeling procedures: whether it is informative depends on the labeling methods themselves producing relevant exemplar prompts. 

We report the average activation values of positive and neutral exemplars for \maia, \milan, and human labels across all tested models in Figure~\ref{fig:eval-results}. \textbf{\maia outperforms \milan across all models and is often on par with expert predictions.} This trend persists across different averaging techniques (such as normalizing by activation percentile, see Appendix \ref{app:normalized_eval}). While \milan is a relevant neuron labeling baseline, we note that comparisons to task-specific procedures that use learned models to label a fixed set of exemplars only evaluate part of \maia's full functionality. \maia is easily adaptable to downstream auditing applications that require additional experimentation, where one-shot neuron labeling procedures are insufficient (see Section \ref{sec: spurious}). Table~\ref{tab:layer-eval-appendix} provides additional comparisons of \maia to neuron labeling baselines, and shows evaluation results by layer. 

\subsection{Synthetic neurons}
\label{sec:syn_eval}

\begin{figure}[t]
    \centering
    \includegraphics[width=\linewidth]{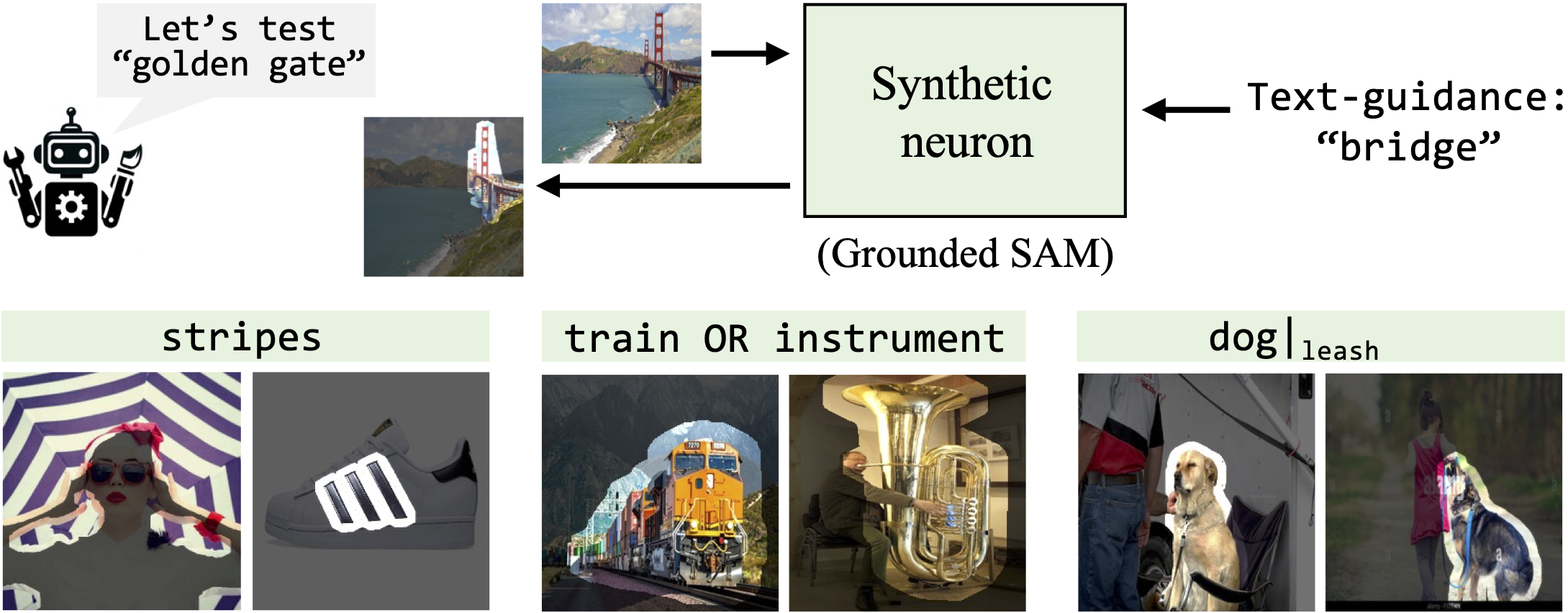}
    \caption{\textbf{Synthetic neuron implementation.} Segmentation of input images is performed by an open-set concept detector with text guidance specifying ground-truth neuron selectivity. Synthetic neurons return masked images and synthetic activation values corresponding to the probability a concept is present in the image.} 
    \vspace{-4mm}
    \label{fig:synthetic_schematic}
\end{figure}

\begin{figure}[t]
    \centering
    \includegraphics[width=\linewidth]{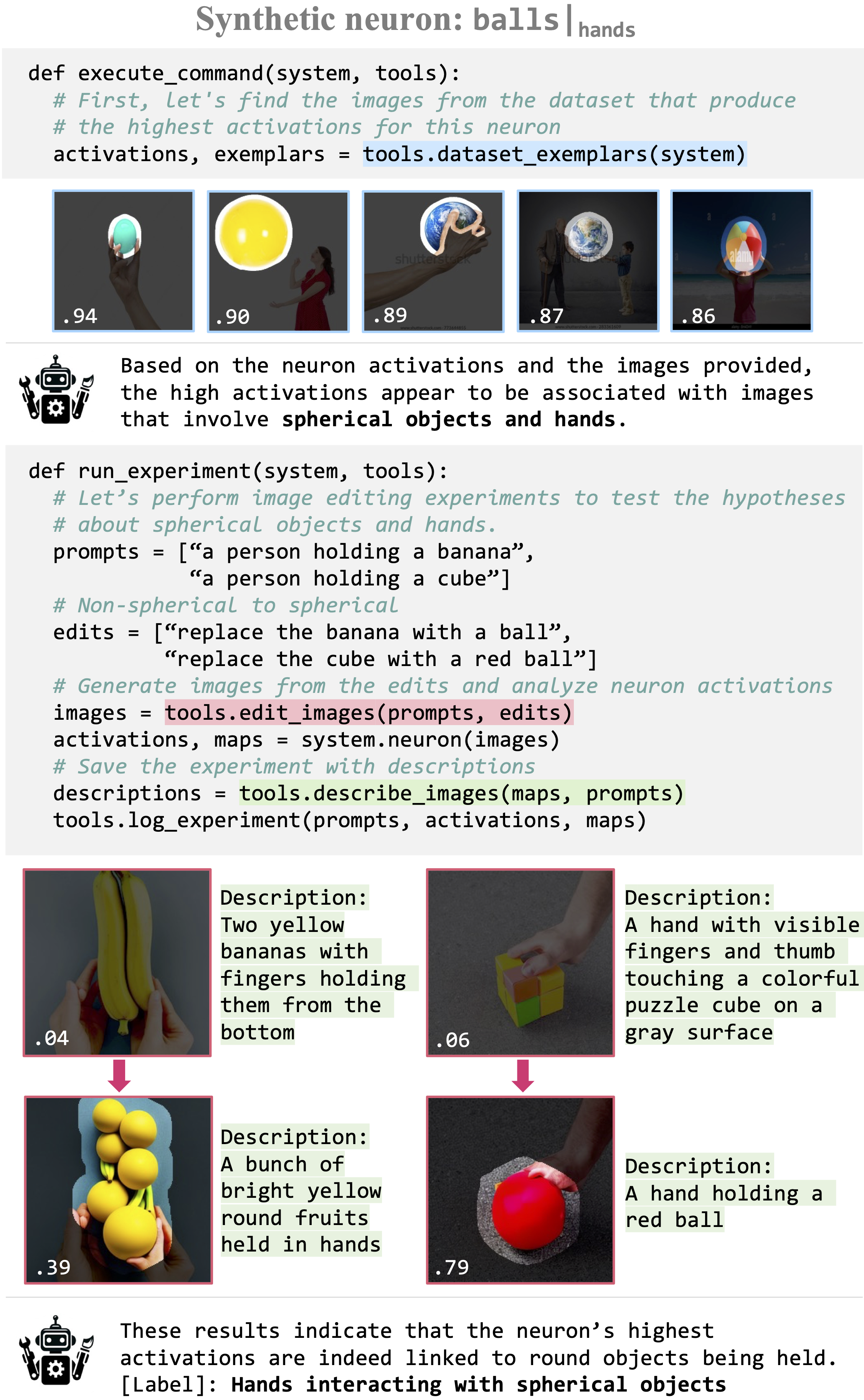}
    \vspace{-4mm}
    \caption{\textbf{\maia synthetic neuron interpretation.} 
    }
    \vspace{-5mm}
    \label{fig:synthetic_interp}
\end{figure}

Following the procedure in \citet{schwettmann2023find} for validating the performance of automated interpretability methods on synthetic test systems mimicking real-world behaviors, we construct a set of \textit{synthetic vision neurons} with known ground-truth selectivity. We simulate concept detection performed by neurons inside vision models using semantic segmentation. Synthetic neurons are built using an open-set concept detector that combines Grounded DINO \citep{liu2023grounding} with SAM \citep{kirillov2023segany} to perform text-guided image segmentation. The ground-truth behavior of each neuron is determined by a text description of the concept(s) the neuron is selective for (Figure \ref{fig:synthetic_schematic}). To capture real-world behaviors, we derive neuron labels from \milannotations, a dataset of 60K human annotations of neurons across seven trained vision models \citep{hernandez2022natural}. Neurons in the wild display a diversity of behaviors: some respond to individual concepts, while others respond to complex combinations of concepts \citep{netdissect2017, fong2018net2vec, olah2020zoom, mu2021compositional, gurnee2023finding}. We construct three types of synthetic neurons with increasing levels of complexity: monosemantic neurons that recognize single concepts (\eg \textit{stripes}), polysemantic neurons selective for logical disjunctions of concepts (\eg \textit{trains} \verb|OR| \textit{instruments}), and \textit{conditional} neurons that only recognize a concept in presence of another concept (\eg \textit{dog$|_{\text{leash}}$}). Following the instrumentation of real neurons in the \maia API, synthetic vision neurons accept image input and return a masked image highlighting the concept they are selective for (if present), and an activation value (corresponding to the confidence of Grounded DINO in the presence of the concept). Dataset exemplars for synthetic neurons are calculated by computing 15 top-activating images per neuron from the CC3M dataset \cite{sharma2018conceptual}. Figure \ref{fig:synthetic_schematic} shows examples of each type of neuron; the full list of 85 synthetic neurons is provided in Appendix \ref{sec:syn_appendix}. The set of concepts that can be represented by synthetic neurons is limited to simple concepts by the fidelity of open-set concept detection using current text-guided segmentation methods. We verify that all concepts in the synthetic neuron dataset can be segmented by Grounded DINO in combination with SAM, and provide further discussion of the limitations of synthetic neurons in Appendix \ref{sec:syn_appendix}.

MAIA interprets synthetic neurons using the same API and procedure used to interpret neurons in trained vision models (Section \ref{sec:real_eval}). In contrast to neurons in the wild, we can evaluate descriptions of synthetic neurons directly against ground-truth neuron labels. We collect comparative annotations of synthetic neurons from \milan, as well as expert annotators (using the procedure from Section \ref{sec:real_eval} where human experts manually label a subset of 25\% of neurons using the \maia API). We recruit human judges from Amazon Mechanical Turk to evaluate the agreement between synthetic neuron descriptions and ground-truth labels in pairwise two-alternative forced choice (2AFC) tasks. For each task, human judges are shown the ground-truth neuron label (\eg \textit{tail}) and descriptions produced by two labeling procedures (\eg ``fluffy and textured animal tails'' and ``circular objects and animals''), and asked to select which description better matches the ground-truth label. Further details are provided in Appendix \ref{sec:syn_appendix}. Table \ref{tab:syn-eval} shows the results of the 2AFC study (the proportion of trials in which procedure $A$ was favored over $B$, and 95\% confidence intervals). According to human judges, \textbf{\maia labels better agree with ground-truth labels when compared to \milan, and are even slightly preferred over expert labels} on the subset of neurons they described (while human labels are largely preferred over \milan labels). We also use the predictive evaluation framework described in Section \ref{sec:real_eval} to generate positive and neutral sets of exemplar images for all synthetic neurons. Figure \ref{fig:eval-results} shows \maia descriptions are better predictors of synthetic neuron activations than \milan descriptions, on par with labels produced by human experts.

\begin{table}[t!]
\centering
\caption{\textbf{2AFC test.} Human subjects selected which method best agrees with the ground truth synthetic neuron label.
}
\label{tab:syn-eval}
\resizebox{\columnwidth}{!}{
\begin{tabular}{lccc} 
\toprule
 & \maia vs. \milan & \maia vs. Human & Human vs. \milan \\
\midrule
 & $0.73 \pm 4e^{-4}$ & $0.53 \pm 1e^{-3}$  & $0.83 \pm 5e^{-4}$\\
\bottomrule
\vspace{-10mm}
\end{tabular}
}
\end{table}

\subsection{Tool ablation study}
\label{sec:ablation}
\begin{figure}[t]
    \centering
    \includegraphics[width=1\columnwidth]{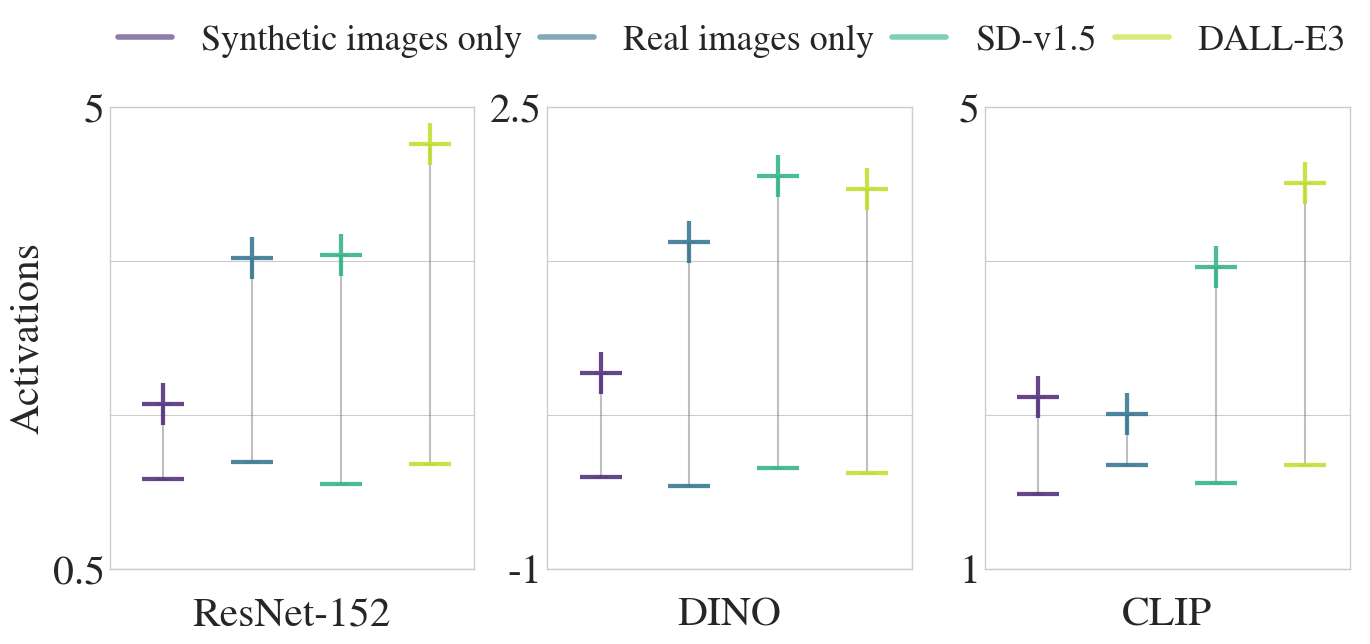}
    \vspace{-6mm}
    \caption{\textbf{Ablation study.} We use the predictive evaluation protocol to quantify \maia's sensitivity to different tools. Top performance is achieved when experimenting with both real and synthetic data, and when using DALL-E 3 for image generation. More details in Appendix~\ref{sec:ablation_appendix}.}
    \vspace{-4mm}
    \label{fig:ablation}
\end{figure}

\maia's modular design enables straightforward addition and removal of tools from its API. We test three different settings to quantify sensitivity to different tools: (i) labeling neurons using only the \code{dataset_exemplar} function without the ability to synthesize images, (ii) relying only on generated inputs without the option to compute maximally activating dataset exemplars, and (iii) replacing the Stable Diffusion \code{text2image} backbone with DALL-E 3. While the first two settings do not fully compromise performance, neither ablated API achieves the same average accuracy as the full \maia system (Figure \ref{fig:ablation}). These results emphasize the combined utility of tools for experimenting with real-world and synthetic inputs: \maia performs best when initializing experiments with dataset exemplars and running additional tests with synthetic images. Methods like \milan that label precomputed exemplars  could thus be incorporated into the \maia API as tools, and used to initialize experimentation. We also find that using DALL-E as the {\small\texttt{text2image}} backbone improves performance (Figure \ref{fig:ablation}). This suggests that the agent is bounded by the performance of its tools rather than its ability to use them---and as interpretability tools grow in sophistication, so will \maia.

\subsection{MAIA failure modes}
\begin{figure}[t]
    \centering
    \includegraphics[width=0.98\linewidth]{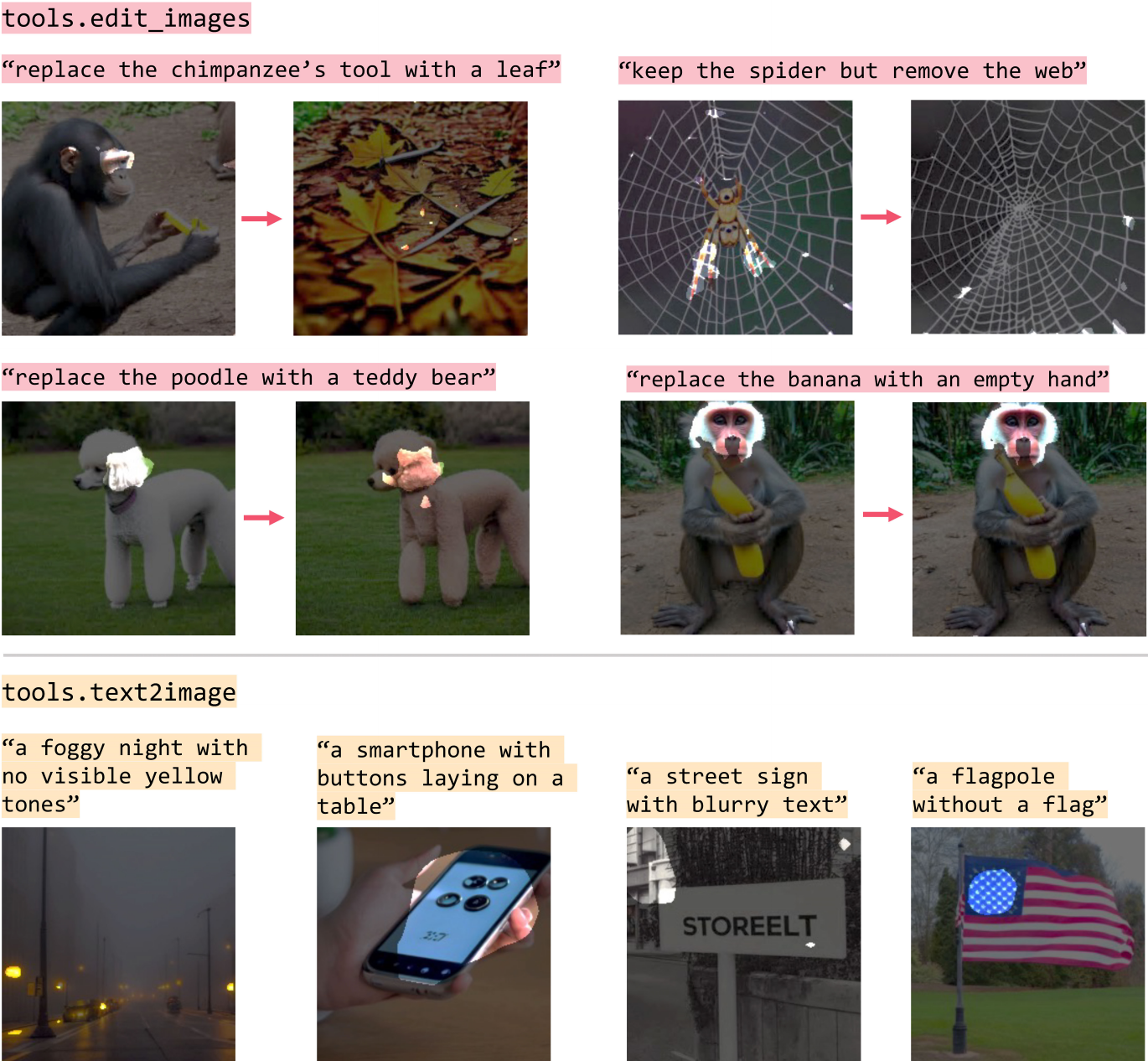}
    \vspace{-4mm}
     \caption{\textbf{\maia tool failures.} \maia is limited by the reliability of its tools. Common image editing failure modes (using InstructPix2Pix) include failing to remove objects, misinterpreting the instructions (e.g. removing the incorrect object), and changing too much or too little of the image. \maia's image generation tool (SD-v1.5) is sometimes unreliable for negative instructions (e.g. a flagpole \textit{without} a flag), and sometimes deviates from the text prompt by adding or excluding image components.}  
    \vspace{-4mm}
    \label{fig:failures_main}
\end{figure}

Consistent with the result in Section \ref{sec:ablation} that \maia performance improves with DALL-E 3, we additionally observe that SD-v1.5 and InstructPix2Pix sometimes fail to faithfully generate and edit images according to \maia's instructions. To mitigate these failures, we instruct \maia to prompt positive image-edits (\eg\textit{replace} the bowtie with a plain shirt) rather than negative edits (\eg\textit{remove} the bowtie), but occasional failures still occur (see Figure \ref{fig:failures_main} and Appendix \ref{sec:failures_appendix}). While proprietary versions of tools may be of higher quality, they also introduce prohibitive rate limits and costs associated with API access. As similar limitations apply to the GPT-4V backbone itself, we tested the performance of free and non-proprietary VLMs as alternative \maia backbones. Currently, off-the-shelf alternatives still significantly lag behind GPT-4V performance (consisitent with evaluation of open-source models' ability to interpret functions in \citet{schwettmann2023find}), but our initial experiments suggest their performance may improve with fine-tuning (see Appendix \ref{app:vlms}). The \maia system is designed modularly so that open-source alternatives can be incorporated in the future as their performance improves.

\vspace{-2mm}
\section{Applications}
\label{sec:experiments}

\maia is a flexible system that automates model understanding tasks at different levels of granularity: from labeling individual features to diagnosing model-level failure modes. To demonstrate the utility of \maia for producing actionable insights for human users \citep{vaughan2020human}, we conduct experiments that apply \maia to two model-level tasks: (i) spurious feature removal and (ii) bias identification in a downstream classification task. In both cases \maia uses the API as described in Section \ref{sec:framework}. In an additional experiment, we evaluate the downstream utility of \maia descriptions by measuring the extent to which they equip humans to make predictions about system behavior (see details in Appendix~\ref{sec:mturk2}).

\vspace{-2mm}
\subsection{Removing spurious features}
\label{sec: spurious}
Learned spurious features impose a challenge when machine learning models are applied in real-world scenarios, where test distributions differ from training set statistics \citep{storkey2009training, beery2018recognition, bissoto2020debiasing, xiao2020noise, singla2021understanding}. We use \maia to remove learned spurious features inside a classification network, finding that \emph{with no access to unbiased examples nor grouping annotations}, \maia can identify and remove such features, improving model robustness under distribution shift by a wide margin, with an accuracy approaching that of fine-tuning on balanced data.

\begin{figure}[t]
    \centering
    \includegraphics[width=\linewidth]{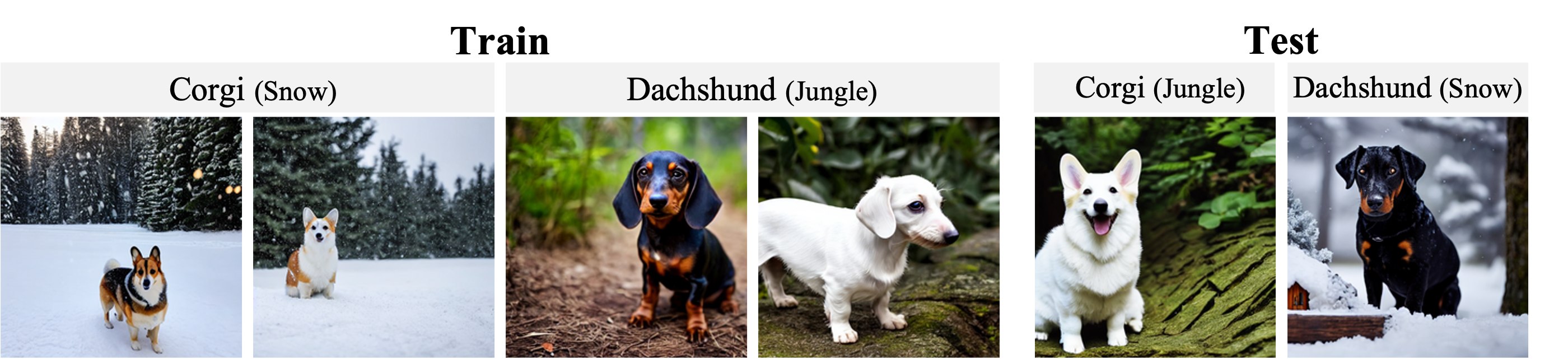}
    \vspace{-8mm}
    \caption{\textbf{Spawrious dataset examples.} Train data contains spurious correlations between dog breeds and their backgrounds.} 
    \label{fig:spawrious}
\end{figure}

We run experiments on ResNet-18 trained on the Spawrious dataset \cite{lynch2023spawrious}, a synthetically generated dataset involving four dog breeds with different backgrounds. In the train set, each breed is spuriously correlated with a certain background type, while in the test set, the breed-background pairings are changed (see Figure \ref{fig:spawrious}). We use \maia to find a subset of final layer neurons that robustly predict a single dog breed independently of spurious features (see Appendix \ref{sec:spaw_examples}). While other methods like~\citet{kirichenko2023layer} remove spurious correlations by retraining the last layer on \emph{balanced} datasets, we only provide \maia access to top-activating images from the \emph{unbalanced} validation set and prompt \maia to run experiments to determine robustness. We then use the features \maia selects to train an unregularized logistic regression model on the unbalanced data.

As a demonstration, we select $50$ of the most informative neurons using $\ell_1$ regularization on the \emph{unbalanced} dataset and have \maia run experiments on each one. \maia selects $22$ neurons it deems to be robust. Traning an unregularized model on this subset significantly improves accuracy, as reported in Table~\ref{tab:spawrious}. 
For comparison, we repeat the same task using interpretability procedures like \milan that rely on precomputed exemplars (both with the original model of \cite{hernandez2022natural} and with GPT-4V, see Appendix~\ref{sec:spaw_experiment_details} for experimental details). Both achieved significantly lower accuracy. To further show that the sparsity of \maia's neuron selection is not the only reason for its performance improvements, we also benchmark \maia's performance against $\ell_1$ regularized fitting on both unbalanced and balanced versions of the dataset.  
On the unbalanced dataset, $\ell_1$ drops in performance when subset size reduces from $50$ to $22$ neurons.
Using a small \emph{balanced} dataset to hyperparameter tune the $\ell_1$ parameter and train the logistic regression model on all neurons achieves performance comparable to $\maia$'s chosen subset, although \maia did not have access to any balanced data. For a fair comparison, we test the performance of an $\ell_1$ model which matches the sparsity of $\maia$, but trained on the balanced dataset.
See Appendix \ref{sec:spaw_experiment_details} for more details. 

\begin{table}[t!]
\centering
\caption{Final layer spurious feature removal results.}
\label{tab:spawrious}
\resizebox{\columnwidth}{!}{
\begin{tabular}{ccccc} 
\toprule
Subset & Selection Method & \# Units & Balanced & Test Acc. \\   
\midrule
All & Original Model & 512 & \ding{55} & 0.731\\
\midrule
\multirow{4}{*}{$\ell_1$ Top 50} & All & 50 & \ding{55} & 0.779 \\
 & Random & 22 & \ding{55} & 0.705 $\pm$ 0.05 \\
 & $\ell_1$ Top 22 & 22 & \ding{55} & 0.757 \\
 & \milan & 23 & \ding{55} & 0.786 \\
 & \milan (GPT-4V) & 23 & \ding{55} & 0.690 \\
 & \maia & 22 & \ding{55} & \textbf{0.837}\\
\midrule
\multirow{2}{*}{All} & $\ell_1$ Hyper. Tuning & 147 & \checkmark & 0.830 \\
 & $\ell_1$ Top 22 & 22 & \checkmark & \textbf{0.865} \\
\bottomrule
\vspace{-2mm}
\end{tabular}
}
\end{table}

\subsection{Revealing biases}
\begin{figure}[t]
    \centering
    \includegraphics[width=\linewidth]{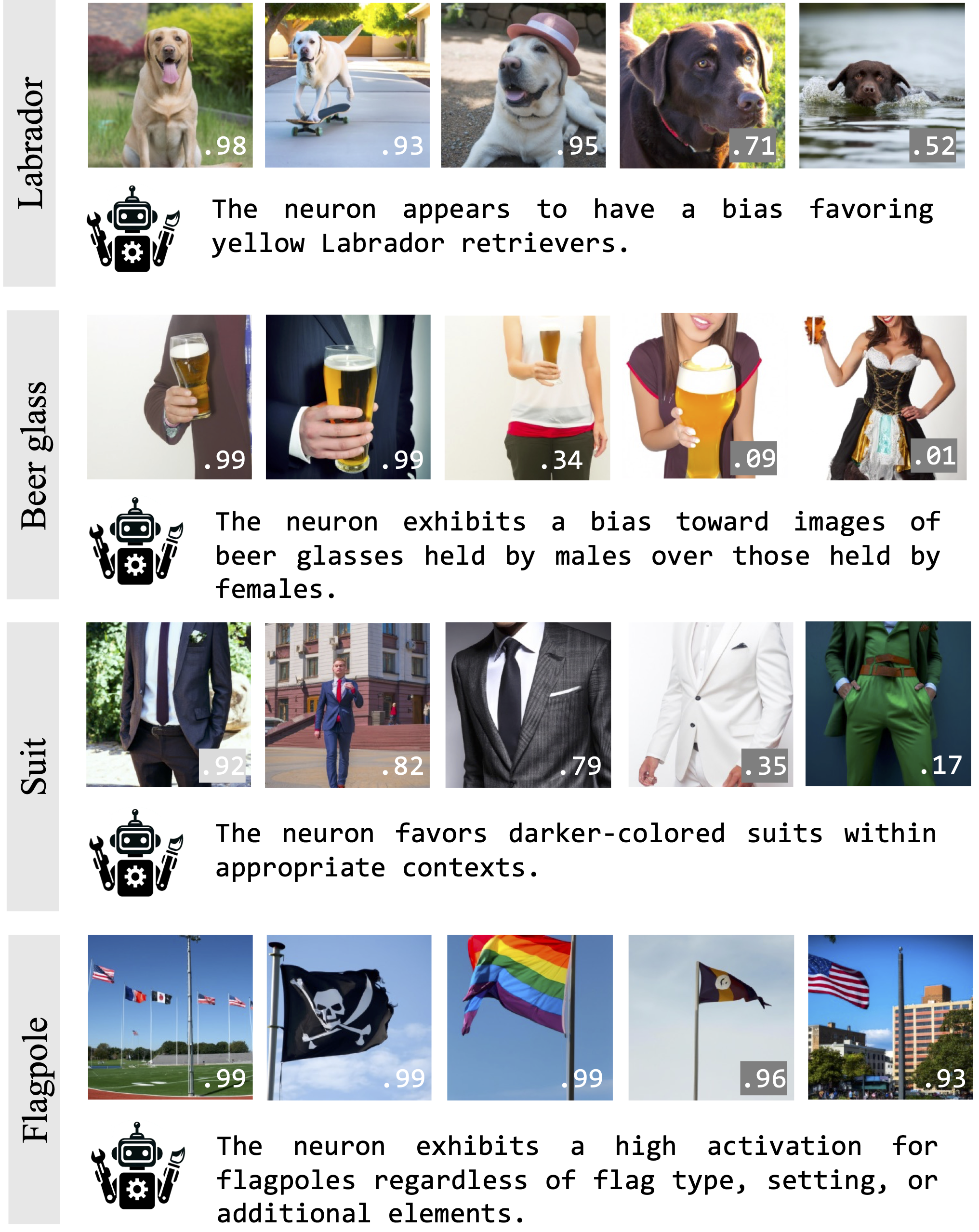}
     \caption{\textbf{\maia bias detection.} \maia iteratively conducts experiments and generates synthetic inputs 
     to surface biases in ResNet-152 output classes. In some cases, \maia discovers uniform behavior over the inputs (\eg \textit{flagpole}).} 
      \vspace{1mm}
    \label{fig:bias_examples}
\end{figure}

\maia can be used to automatically surface model-level biases. Specifically, we apply \maia to investigate biases in the outputs of a CNN (ResNet-152) trained on a supervised ImageNet classification task. The \maia system is easily adaptable to this experiment: the output logit corresponding to a specific class is instrumented using the \code{system} class, and returns class probability for input images. \maia is provided with the class label and instructed (see Appendix \ref{sec:bias-appendix}) to find settings in which the classifier ranks images related to that class with relatively lower probability values, or shows a clear preference for a subset of the class. Figure ~\ref{fig:bias_examples} presents results for a subset of ImageNet classes. This simple paradigm suggests that \maia's generation of synthetic data could be widely useful for identifying regions of the input distribution where a model exhibits poor performance. While this exploratory experiment surfaces only broad failure categories, \maia enables other experiments targeted at end-use cases identifying specific biases.

\section{Conclusion}

We introduce \maia, an agent that automates interpretability tasks including feature interpretation and bias discovery. By composing pretrained modules, \maia conducts experiments to make and test hypotheses about the behavior of other systems. While human supervision is needed to maximize its effectiveness and catch common mistakes, initial experiments with \maia show promise, and we anticipate that interpretability agents will be increasingly useful as they grow in sophistication.

\newpage
\section*{Impact statement}

As AI systems take on higher-stakes roles and become more deeply integrated into research and society, scalable approaches to auditing for reliability will be vital. \maia is a protoype for a tool that can help human users ensure AI systems are transparent, reliable, and equitable.

We think \maia augments, but does not replace, human oversight of AI systems. \maia still requires human supervision to catch mistakes such as confirmation bias and image generation/editing failures. Absence of evidence (from \maia) is not evidence of absence: though \maia's toolkit enables causal interventions on inputs in order to evaluate system behavior, \maia's explanations do not provide formal verification of system performance. 
\section*{Acknowlegements}
We are grateful for the support of the MIT-IBM Watson AI Lab, the Open Philanthropy foundation, Hyundai Motor Company, ARL grant W911NF-18-2-021, Intel, the National Science Foundation under grant CCF-2217064, the Zuckerman STEM Leadership Program, and
the Viterbi Fellowship. The funders had no role in experimental design or analysis,
decision to publish, or preparation of the manuscript. The authors have no competing interests to
report.

\bibliography{main}
\bibliographystyle{icml2024}

\newpage
\appendix
\onecolumn
{
\renewcommand{\thetable}{A\arabic{table}} 

\setcounter{subsection}{0}
\renewcommand{\thesubsection}{A\arabic{subsection}}

\renewcommand{\thefigure}{A\arabic{figure}} 

\setcounter{section}{0}
\setcounter{subsection}{0}
\renewcommand{\thesubsection}{\Alph{section}\arabic{subsection}} 

\textbf{\Large {Appendix}}
\section{\maia Library}
\label{sec:API_appendix}

The full \maia API provided in the system prompt is reproduced below. 

\begin{lstlisting}[style=mystylepython]
import torch
from typing import List, Tuple

class System:
    """
    A Python class containing the vision model and the specific neuron to interact with.
    
    Attributes
    ----------
    neuron_num : int
        The unit number of the neuron.
    layer : string
        The name of the layer where the neuron is located.
    model_name : string
        The name of the vision model.
    model : nn.Module
        The loaded PyTorch model.

    Methods
    -------
    load_model(model_name: str) -> nn.Module
        Gets the model name and returns the vision model from PyTorch library.
    neuron(image_list: List[torch.Tensor]) -> Tuple[List[int], List[str]]
        returns the neuron activation for each image in the input image_list as well as the activation map 
        of the neuron over that image, that highlights the regions of the image where the activations 
        are higher (encoded into a Base64 string).
    """
    def __init__(self, neuron_num: int, layer: str, model_name: str, device: str):
        """
        Initializes a neuron object by specifying its number and layer location and the vision model that the neuron belongs to.
        Parameters
        -------
        neuron_num : int
            The unit number of the neuron.
        layer : str
            The name of the layer where the neuron is located.
        model_name : str
            The name of the vision model that the neuron is part of.
        device : str
            The computational device ('cpu' or 'cuda').
        """
        self.neuron_num = neuron_num
        self.layer = layer
        self.device = torch.device(f"cuda:{device}" if torch.cuda.is_available() else "cpu")       
        self.model = self.load_model(model_name)


    def load_model(self, model_name: str) -> torch.nn.Module:
        """
        Gets the model name and returns the vision model from pythorch library.
        Parameters
        ----------
        model_name : str
            The name of the model to load.
        
        Returns
        -------
        nn.Module
            The loaded PyTorch vision model.
        
        Examples
        --------
        >>> # load "resnet152"
        >>> def run_experiment(model_name) -> nn.Module:
        >>>   model = load_model(model_name: str)
        >>>   return model
        """
        return load_model(model_name)
    
    def neuron(self, image_list: List[torch.Tensor]) -> Tuple[List[int], List[str]]:
        """
        The function returns the neuron's maximum activation value (in int format) for each of the images in the list as well as the activation map of the neuron over each of the images that highlights the regions of the image where the activations are higher (encoded into a Base64 string).
        
        Parameters
        ----------
        image_list : List[torch.Tensor]
            The input image
        
        Returns
        -------
        Tuple[List[int], List[str]]
            For each image in image_list returns the activation value of the neuron on that image, and a masked image, 
            with the region of the image that caused the high activation values highlighted (and the rest of the image is darkened). Each image is encoded into a Base64 string.

        
        Examples
        --------
        >>> # test the activation value of the neuron for the prompt "a dog standing on the grass"
        >>> def run_experiment(system, tools) -> Tuple[int, str]:
        >>>     prompt = ["a dog standing on the grass"]
        >>>     image = tools.text2image(prompt)
        >>>     activation_list, activation_map_list = system.neuron(image)
        >>>     return activation_list, activation_map_list
        >>> # test the activation value of the neuron for the prompt "a dog standing on the grass" and the neuron activation value for the same image but with a lion instead of a dog
        >>> def run_experiment(system, tools) -> Tuple[int, str]:
        >>>     prompt = ["a dog standing on the grass"]
        >>>     edits = ["replace the dog with a lion"]
        >>>     all_image, all_prompts = tools.edit_images(prompt, edits)
        >>>     activation_list, activation_map_list = system.neuron(all_images)
        >>>     return activation_list, activation_map_list
        
        """
        return neuron(image_list)

class Tools:
    """
    A Python class containing tools to interact with the neuron implemented in the system class, 
    in order to run experiments on it.
    
    Attributes
    ----------
    experiment_log: str
        A log of all the experiments, including the code and the output from the neuron.


    Methods
    -------
    dataset_exemplars(system: object) -> Tuple(List[int],List[str])
        This experiment provides good coverage of the behavior observed on a very large dataset of images and therefore represents the typical behavior of the neuron on real images.
        This function characterizes the prototipycal behavior of the neuron by computing its activation on all images in the ImageNet dataset and returning the 15 highest activation values and the images that produced them. 
        The images are masked to highlight the specific regions that produce the maximal activation. The images are overlaid with a semi-opaque mask, such that the maximally activating regions remain unmasked.
    edit_images(prompt_list_org_image : List[str], editing_instructions_list : List[str]) -> Tuple[List[Image.Image], List[str]]
        This function enables loclized testing of specific hypotheses about how variations on the content of a single image affect neuron activations.
        Gets a list of input prompt and a list of corresponding editing instructions, then generate images according to the input prompts and edits each image based on the instructions given in the prompt using a text-based image editing model.
        This function is very useful for testing the causality of the neuron in a controlled way, for example by testing how the neuron activation is affected by changing one aspect of the image.
        IMPORTANT: Do not use negative terminology such as "remove ...", try to use terminology like "replace ... with ..." or "change the color of ... to ...".
    text2image(prompt_list: str) -> Tuple[torcu.Tensor]
        Gets a list of text prompt as an input and generates an image for each prompt in the list using a text to image model.
        The function returns a list of images.
    summarize_images(self, image_list: List[str]) -> str:    
        This function is useful to summarize the mutual visual concept that appears in a set of images.
        It gets a list of images at input and describes what is common to all of them, focusing specifically on unmasked regions.
    describe_images(synthetic_image_list: List[str], synthetic_image_title:List[str]) -> str
        Provides impartial descriptions of images. Do not use this function on dataset exemplars.
        Gets a list of images and generat a textual description of the semantic content of the unmasked regions within each of them.
        The function is blind to the current hypotheses list and therefore provides an unbiased description of the visual content.
    log_experiment(activation_list: List[int], image_list: List[str], image_titles: List[str], image_textual_information: Union[str, List[str]]) -> None
        documents the current experiment results as an entry in the experiment log list. if self.activation_threshold was updated by the dataset_exemplars function, 
        the experiment log will contains instruction to continue with experiments if activations are lower than activation_threshold.
        Results that are loged will be available for future experiment (unlogged results will be unavailable).
        The function also update the attribure "result_list", such that each element in the result_list is a dictionary of the format: {"<prompt>": {"activation": act, "image": image}}
        so the list contains all the resilts that were logged so far.
    """

    def __init__(self):
        """
        Initializes the Tools object.

        Parameters
        ----------
        experiment_log: store all the experimental results
        """
        self.experiment_log = []
        self.results_list = []


    def dataset_exemplars(self, system: object) -> Tuple(List[int],List[str])
        """
        This method finds images from the ImageNet dataset that produce the highest activation values for a specific neuron.
        It returns both the activation values and the corresponding exemplar images that were used 
        to generate these activations (with the highly activating region highlighted and the rest of the image darkened). 
        The neuron and layer are specified through a 'system' object.
        This experiment is performed on real images and will provide a good approximation of the neuron behavior.
        
        Parameters
        ----------
        system : object
            An object representing the specific neuron and layer within the neural network.
            The 'system' object should have 'layer' and 'neuron_num' attributes, so the dataset_exemplars function 
            can return the exemplar activations and masked images for that specific neuron.

        Returns
        -------
        tuple
            A tuple containing two elements:
            - The first element is a list of activation values for the specified neuron.
            - The second element is a list of exemplar images (as Base64 encoded strings) corresponding to these activations.

        Example
        -------
        >>> def run_experiment(system, tools)
        >>>     activation_list, image_list = self.dataset_exemplars(system)
        >>>     return activation_list, image_list
        """
        
        return dataset_exemplars(system)

    def edit_images(self, prompt_list_org_image : List[str], editing_instructions_list : List[str]) -> Tuple[List[Image.Image], List[str]]:
        """
        This function enables localized testing of specific hypotheses about how variations in the content of a single image affect neuron activations.
        Gets a list of prompts to generate images, and a list of corresponding editing instructions as inputs. Then generates images based on the image prompts and edits each image based on the instructions given in the prompt using a text-based image editing model (so there is no need to generate the images outside of this function).
        This function is very useful for testing the causality of the neuron in a controlled way, for example by testing how the neuron activation is affected by changing one aspect of the image.
        IMPORTANT: for the editing instructions, do not use negative terminology such as "remove ...", try to use terminology like "replace ... with ..." or "change the color of ... to"
        The function returns a list of images, constructed in pairs of original images and their edited versions, and a list of all the corresponding image prompts and editing prompts in the same order as the images.

        Parameters
        ----------
        prompt_list_org_image : List[str]
            A list of input prompts for image generation. These prompts are used to generate images which are to be edited by the prompts in editing_instructions_list.
        editing_instructions_list : List[str]
            A list of instructions for how to edit the images in image_list. Should be the same length as prompt_list_org_image.
            Edits should be relatively simple and describe replacements to make in the image, not deletions.

        Returns
        -------
        Tuple[List[Image.Image], List[str]]
            A list of all images where each unedited image is followed by its edited version. 
            And a list of all the prompts corresponding to each image (e.g. the input prompt followed by the editing instruction).

        Examples
        --------
        >>> # test the activation value of the neuron for the prompt "a dog standing on the grass" and the neuron activation value for the same image but with a cat instead of a dog
        >>> def run_experiment(system, tools) -> Tuple[int, str]:
        >>>     prompt = ["a dog standing on the grass"]
        >>>     edits = ["replace the dog with a cat"]
        >>>     all_image, all_prompts = tools.edit_images(prompt, edits)
        >>>     activation_list, activation_map_list = system.neuron(all_images)
        >>>     return activation_list, activation_map_list
        >>> # test the activation value of the neuron for the prompt "a dog standing on the grass" and the neuron activation values for the same image but with a different action instead of "standing":
        >>> def run_experiment(system, tools) -> Tuple[int, str]:
        >>>     prompts = ["a dog standing on the grass"]*3
        >>>     edits = ["make the dog sit","make the dog run","make the dog eat"]
        >>>     all_images, all_prompts = tools.edit_images(prompts, edits)
        >>>     activation_list, activation_map_list = system.neuron(all_images)
        >>>     return activation_list, activation_map_list
        """

        return edit_images(image, edits)


    def text2image(self, prompt_list: List[str]) -> List[Image.Image]:
        """Gets a list of text prompts as input, generates an image for each prompt in the list using a text to image model.
        The function returns a list of images.

        Parameters
        ----------
        prompt_list : List[str]
            A list of text prompts for image generation.

        Returns
        -------
        List[Image.Image]
            A list of images, corresponding to each of the input prompts. 

        Examples
        --------
        >>> # test the activation value of the neuron for the prompt "a dog standing on the grass"
        >>> def run_experiment(system, tools) -> Tuple[int, str]:
        >>>     prompt = ["a dog standing on the grass"]
        >>>     image = tools.text2image(prompt)
        >>>     activation_list, activation_map_list = system.neuron(image)
        >>>     return activation_list, activation_map_list
        >>> # test the activation value of the neuron for the prompt "a fox and a rabbit watch a movie under a starry night sky" "a fox and a bear watch a movie under a starry night sky" "a fox and a rabbit watch a movie at sunrise"
        >>> def run_experiment(system, tools) -> Tuple[int, str]:
        >>>     prompt_list = ["a fox and a rabbit watch a movie under a starry night sky", "a fox and a bear watch a movie under a starry night sky","a fox and a rabbit watch a movie at sunrise"]
        >>>     images = tools.text2image(prompt_list)
        >>>     activation_list, activation_map_list = system.neuron(images)
        >>>     return activation_list, activation_map_list
        """

        return text2image(prompt_list)

    def summarize_images(self, image_list: List[str]) -> str:
        """
        This function is useful to summarize the mutual visual concept that appears in a set of images.
        It gets a list of images at input and describes what is common to all of them, focusing specifically on unmasked regions.

        Parameters
        ----------
        image_list : list
            A list of images in Base64 encoded string format.
        
        Returns
        -------
        str
            A string with a descriptions of what is common to all the images.

        Example
        -------
        >>> # tests dataset exemplars and return textual summarization of what is common for all the maximally activating images
        >>> def run_experiment(system, tools):
        >>>     activation_list, image_list = self.dataset_exemplars(system)
        >>>     prompt_list = []
        >>>     for i in range(len(activation_list)):
        >>>          prompt_list.append(f'dataset exemplar {i}') # for the dataset exemplars we don't have prompts, therefore need to provide text titles
        >>>     summarization = tools.summarize_images(image_list)
        >>>     return summarization
        """

        return summarize_images(image_list)

    def describe_images(self, image_list: List[str], image_title:List[str]) -> str:
        """
        Provides impartial description of the highlighted image regions within an image.
        Generates textual descriptions for a list of images, focusing specifically on highlighted regions.
        This function translates the visual content of the highlited region in the image to a text description. 
        The function operates independently of the current hypothesis list and thus offers an impartial description of the visual content.        
        It iterates through a list of images, requesting a description for the 
        highlighted (unmasked) regions in each synthetic image. The final descriptions are concatenated 
        and returned as a single string, with each description associated with the corresponding 
        image title.

        Parameters
        ----------
        image_list : list
            A list of images in Base64 encoded string format.
        image_title : callable
            A list of strings with the image titles that will be use to list the different images. Should be the same length as image_list. 

        Returns
        -------
        str
            A concatenated string of descriptions for each image, where each description 
            is associated with the image's title and focuses on the highlighted regions 
            in the image.

        Example
        -------
        >>> def run_experiment(system, tools):
        >>>     prompt_list = ["a fox and a rabbit watch a movie under a starry night sky", "a fox and a bear watch a movie under a starry night sky","a fox and a rabbit watch a movie at sunrise"]
        >>>     images = tools.text2image(prompt_list)
        >>>     activation_list, image_list = system.neuron(images)
        >>>     descriptions = tools.describe_images(image_list, prompt_list)
        >>>     return descriptions
        """

        return describe_images(image_list, image_title)


    def log_experiment(self, activation_list: List[int], image_list: List[str], image_titles: List[str], image_textual_information: Union[str, List[str]]):
        """documents the current experiment results as an entry in the experiment log list. if self.activation_threshold was updated by the dataset_exemplars function, 
        the experiment log will contain instruction to continue with experiments if activations are lower than activation_threshold.
        Results that are logged will be available for future experiments (unlogged results will be unavailable).
        The function also updates the attribute "result_list", such that each element in the result_list is a dictionary of the format: {"<prompt>": {"activation": act, "image": image}}
        so the list contains all the results that were logged so far.

        Parameters
        ----------
        activation_list : List[int]
            A list of the activation values that were achived for each of the images in "image_list". 
        image_list : List[str]
            A list of the images that were generated using the text2image model and were tested. Should be the same length as activation_list. 
        image_titles : List[str]
            A list of the text lables for the images. Should be the same length as activation_list. 
        image_textual_information: Union[str, List[str]]
            A string or a list of strings with additional information to log such as the image summarization and/or the image textual descriptions.
        
        Returns
        -------
            None
            

        Examples
        --------
        >>> # tests the activation value of the neuron for the prompts "a fox and a rabbit watch a movie under a starry night sky", "a fox and a bear watch a movie under a starry night sky", "a fox and a rabbit watch a movie at sunrise", describes the images and logs the results and the image descriptions 
        >>> def run_experiment(system, tools):
        >>>     prompt_list = ["a fox and a rabbit watch a movie under a starry night sky", "a fox and a bear watch a movie under a starry night sky","a fox and a rabbit watch a movie at sunrise"]
        >>>     images = tools.text2image(prompt_list)
        >>>     activation_list, activation_map_list = system.neuron(images)
        >>>     descriptions = tools.describe_images(images, prompt_list)
        >>>     tools.log_experiment(activation_list, activation_map_list, prompt_list, descriptions)
        >>>     return 
        >>> # tests dataset exemplars, use umage summarizer and logs the results
        >>> def run_experiment(system, tools):
        >>>     activation_list, image_list = self.dataset_exemplars(system)
        >>>     prompt_list = []
        >>>     for i in range(len(activation_list)):
        >>>          prompt_list.append(f'dataset_exemplars {i}') # for the dataset exemplars we don't have prompts, therefore need to provide text titles
        >>>     summarization = tools.summarize_images(image_list)
        >>>     log_experiment(activation_list, activation_map_list, prompt_list, summarization)
        >>>     return
        >>> # test the effect of changing a dog into a cat. Describes the images and logs the results.
        >>> def run_experiment(system, tools) -> Tuple[int, str]:
        >>>     prompt = ["a dog standing on the grass"]
        >>>     edits = ["replace the dog with a cat"]
        >>>     all_images, all_prompts = tools.edit_images(prompt, edits)
        >>>     activation_list, activation_map_list = system.neuron(all_images)
        >>>     descriptions = tools.describe_images(activation_map_list, all_prompts)
        >>>     tools.log_experiment(activation_list, activation_map_list, all_prompts, descriptions)
        >>>     return 
        >>> # test the effect of changing the dog's action on the activation values. Describes the images and logs the results.
        >>> def run_experiment(system, prompt_list) -> Tuple[int, str]:
        >>>     prompts = ["a dog standing on the grass"]*3
        >>>     edits = ["make the dog sit","make the dog run","make the dog eat"]
        >>>     all_images, all_prompts = tools.edit_images(prompts, edits)
        >>>     activation_list, activation_map_list = system.neuron(all_images)
        >>>     descriptions = tools.describe_images(activation_map_list, all_prompts)
        >>>     tools.log_experiment(activation_list, activation_map_list, all_prompts, descriptions)
        >>>     return 
        """
        
        return log_experiment(activation_list, image_list, prompt_list, description)
\end{lstlisting}

\section{\maia user prompt: neuron description}
\label{sec:user_appendix}

\begin{mygraybox}{}
Your overall task is to describe the visual concepts that maximally activate a neuron inside a deep network for computer vision. 
To do that you are provided with a library of Python functions to run experiments on the specific neuron 
(inside the "System" class) given the functions provided in the "Tools" class. Make sure to use a variety of tools from the library to maximize your experimentation power.
Some neurons might be selective for very specific concepts, a group of unrelated concepts, or a general concept, so try to be creative in your experiment and try to test both general and specific concepts. If a neuron is selective for multiple concepts, you should describe each of those concepts in your final description. 
At each experiment step, write Python code that will conduct your experiment on the tested neuron, using the following format:
[CODE]: 
```python
def run_experiment(system, tools)
    # gets an object of the system class, an object of the tool class, and performs experiments on the neuron with the tools
    ...
    tools.log_experiment(...)
```
Finish each experiment by documenting it by calling the "log_experiment" function. Do not include any additional implementation other than this function. Do not call "execute_command" after defining it. Include only a single instance of experiment implementation at each step.

Each time you get the output of the neuron, try to summarize what the inputs that activate the neuron have in common (where that description is not influenced by previous hypotheses). Then, write multiple hypotheses that 
could explain the visual concept(s) that activate the neuron. Note that the neuron can be selective for more than one concept.
For example, these hypotheses could list multiple concepts that the neuron is selective for (e.g. dogs OR cars 
OR birds), provide different explanations for the same concept, describe the same concept at different levels 
of abstraction, etc. Some of the concepts can be quite specific, test hypotheses that are both general and very specific.
Then write a list of initial hypotheses about the neuron selectivity in the format:
[HYPOTHESIS LIST]: 
Hypothesis_1: <hypothesis_1>
...
Hypothesis_n: <hypothesis_n>.

After each experiment, wait to observe the outputs of the neuron. Then your goal is to draw conclusions from the data, update your list of hypotheses, and write additional experiments to test them. Test the effects of both local and global differences in images using the different tools in the library. 
If you are unsure about the results of the previous experiment you can also rerun it, or rerun a modified version of it with additional tools. Use the following format:
[HYPOTHESIS LIST]: ## update your hypothesis list according to the image content and related activation values. 
Only update your hypotheses if image activation values are higher than previous experiments.
[CODE]: ## conduct additional experiments using the provided python library to test *ALL* the hypotheses. Test 
different and specific aspects of each hypothesis using all of the tools in the library. Write code to run the experiment in the same format provided above. Include only a single instance of experiment implementation.

Continue running experiments until you prove or disprove all of your hypotheses. Only when you are confident in your hypothesis after proving it in multiple experiments, output your final description of the neuron in the following format:

[DESCRIPTION]: <final description> ## Your description should be selective (e.g. very specific: "dogs running on the grass" and not just "dog") and complete (e.g. include all relevant aspects the neuron is selective for). 
In cases where the neuron is selective for more than one concept, include in your description a list of all the concepts separated by logical "OR".

[LABEL]: <final label drived from the hypothesis or hypotheses> ## a label for the neuron generated from the hypothesis (or hypotheses) you are most confident in after running all experiments. They should be concise and 
complete, for example, "grass surrounding animals", "curved rims of cylindrical objects", "text displayed on computer screens", "the blue sky background behind a bridge", and "wheels on cars" are all appropriate. You should capture the concept(s) the neuron is selective for. Only list multiple hypotheses if the neuron is selective for multiple distinct concepts. List your hypotheses in the format:
[LABEL 1]: <label 1>
[LABEL 2]: <label 2>

\end{mygraybox}

\newpage
\section{Evaluation experiment details}
\label{sec:eval_appendix}
In Table \ref{tab:layer-eval-appendix} we provide full evaluation results by layer, as well as the number of units evaluated in each layer. Units were sampled uniformly at random, for larger numbers of units in later layers with more interpretable features.

\begin{table}[h]
\centering
\small
\vspace{-.3cm}
\caption{Evaluation results by layer}
\begin{tabular}{llccccccccc} 
\toprule
\multicolumn{3}{c}{} & \multicolumn{2}{c}{\milan} & \multicolumn{2}{c}{CLIP-Dissect} & \multicolumn{2}{c}{\maia} & \multicolumn{2}{c}{Human} \\
\cmidrule(lr){4-5} \cmidrule(lr){6-7} \cmidrule(lr){8-9} 
Arch. & Layer & \# Units & $+$ & $-$ & $+$ & $-$ & $+$  & $-$  & $+$  & $-$  \\ 
\midrule
\multirow{5}{*}{ResNet-152} & conv. 1 & 10 & 7.23 & 3.38 & 6.7 & 2.71 & 7.28       & 3.53       & 7.83        & 3.16       \\ 
                        & res. 1  & 15 & 0.82  & 0.73 & 0.83 &  0.55 & 0.78       & 0.69       & 0.46        & 0.64       \\ 
                        & res. 2  & 20 & 0.98 & 0.92 & 1.02 & 0.66 & 1.02       & 0.90       & 0.83        & 0.95       \\ 
                        & res. 3  & 25 & 1.28  & 0.72 & 0.98 & 0.68 & 1.28       & 0.70       & 2.59        & 0.58       \\ 
                        & res. 4  & 30 & 5.41  & 2.04 & 7.1 & 1.61 & 7.10       & 1.74       & 7.89        & 1.99       \\ 
                        &\textbf{ Avg. } & & \textbf{2.99} & \textbf{1.42} & \textbf{3.37}& \textbf{1.14}& \textbf{3.50} & \textbf{1.33} & \textbf{4.15} & \textbf{1.34} \\
                        \midrule
\multirow{7}{*}{DINO-ViT} & MLP 1  & 5 & 1.10       & 0.94    & 0.76 & 0.94    & 1.19       & 0.74       & 0.63        & 0.34        \\ 
                      & MLP 3  & 5 & 0.63       & 0.96    & 0.96 & 1.06    & 0.81       & 0.87       & 0.55        & 0.89        \\ 
                      & MLP 5  & 20 & 0.85       & 1.01   & 1.11 & 0.94      & 1.33       & 0.97       & 0.84        & 0.84        \\ 
                      & MLP 7  & 20 & 1.42       & 0.77   & 1.16 & 0.66     & 1.67       & 0.82       & 2.58        & 0.54        \\ 
                      & MLP 9  & 25 & 3.50       & -1.15  & 0.81 & -0.83    & 6.31       & -0.81      & 8.64        & -1.06       \\ 
                      & MLP 11 & 25 & -1.56      & -1.94  & -1.12 & -1.65    & -1.41      & -1.84      & -0.61       & -2.49       \\ 
                      & \textbf{Avg.} & & \textbf{1.03} & \textbf{-0.32} & \textbf{0.44} & \textbf{-0.2} & \textbf{1.93} & \textbf{0.54} & \textbf{1.97} & \textbf{-0.23} \\ 
\midrule
\multirow{5}{*}{CLIP-RN50} & res. 1 & 10 & 1.92       & 2.16  & 2.34 & 1.82     & 2.10       & 2.07       & 1.65        & 2.15        \\ 
                      & res. 2 & 20 & 2.54       & 2.46       & 2.61 & 1.91 & 2.78       & 2.39       & 2.22        & 2.81        \\ 
                      & res. 3 & 30 & 2.24       & 1.70       & 2.33 & 1.45 & 2.27       & 1.70       & 2.41        & 1.96        \\ 
                      & res. 4 & 40 & 3.56       & 1.30       & 4.75 & 1.36 & 4.90       & 1.39       & 4.92        & 1.29        \\ 
                      & \textbf{Avg.} &  & \textbf{2.79} & \textbf{1.74} & \textbf{3.36} & \textbf{1.55} & \textbf{3.41} & \textbf{1.75} & \textbf{3.29} & \textbf{1.89} \\ 
                      \bottomrule
\vspace{-5mm}
\label{tab:layer-eval-appendix}
\end{tabular}
\end{table}
\subsection{Predictive evaluation with activation normalizing.}
\label{app:normalized_eval}
Averaging the raw activation values might be largely affected by neurons with a relatively high activation range. Optimally activation values should be reported in percentile, relative to the activation range of each neuron individually, however neuron activation ranges are difficult to be precisely estimated. Instead, we perform the following estimation: 
we normalized the activation value of each neuron by its 95\% percentile activation value (assuming 0\% corresponds to an activation value of 0, which is a valid assumption for convolutional networks with ReLU activation like ResNet and CLIP). Results are reported in Table \ref{tab:normalized_eval}, showing the same trends as pre-normalization. 

\begin{table}[h!]
\vspace{-.5cm}
\label{tab:normalized_eval}
    \centering
    \small
    \caption{Predictive evaluation results normalizing by 95\% percentile}
    \begin{tabular}{c|cccc|cccc}
        \hline
        & \multicolumn{4}{c}{ResNet-152} & \multicolumn{4}{c}{CLIP-RN5} \\
        \hline
        exemplars &\milan & CLIP-Dissect & \maia & Human & \milan & CLIP-Dissect & \maia & Human \\
        \hline
        + & 1.51 & 1.55 & 1.63 & 2.34 & 1.53 & 1.89 & 1.9 & 1.4 \\
        - & 0.83 & 0.75 & 0.78 & 0.63 & 0.77 & 0.77 & 0.76 & 0.75 \\
        \hline
    \end{tabular}
\end{table}

\subsection{Ablation studies}
\label{sec:ablation_appendix}
We use the subset of 25\% neurons labeled by human experts to perform the ablation studies. Results of the predictive evaluation procedure described in Section \ref{sec:evaluation} are shown below. Using DALL-E 3 improves performance over SD-v1.5.

\begin{minipage}{\linewidth}
\centering%
\small
\vspace{-3mm}
\tabcaption{Numerical data for the ablations in Figure \ref{fig:ablation}.}%
\label{tab:ablation-eval}%

\begin{tabular}{lcccc} 
\toprule
& & ImageNet & SD-v1.5 & DALL-E 3 \\
\midrule
ResNet-152  & $+$ & 3.53 & 3.56 & 4.64 \\
            & $-$ & 1.54 & 1.33 & 1.53 \\
\midrule
DINO-ViT    & $+$ & 1.48 & 1.98 & 1.88 \\
            & $-$ & -0.37 & -0.23 & -0.27  \\
\midrule
CLIP-RN50   & $+$ & 2.34 & 3.62 & 4.34 \\
            & $-$ & 1.90 & 1.75 & 1.90 \\
\bottomrule
\end{tabular}

\end{minipage}
\\[\intextsep]

\newpage
\subsection{Human expert neuron description using the \maia tool library}
\label{sec:human_eval}

\begin{wrapfigure}{r}{0.6\textwidth}
  \centering
  \includegraphics[width=0.95\linewidth]{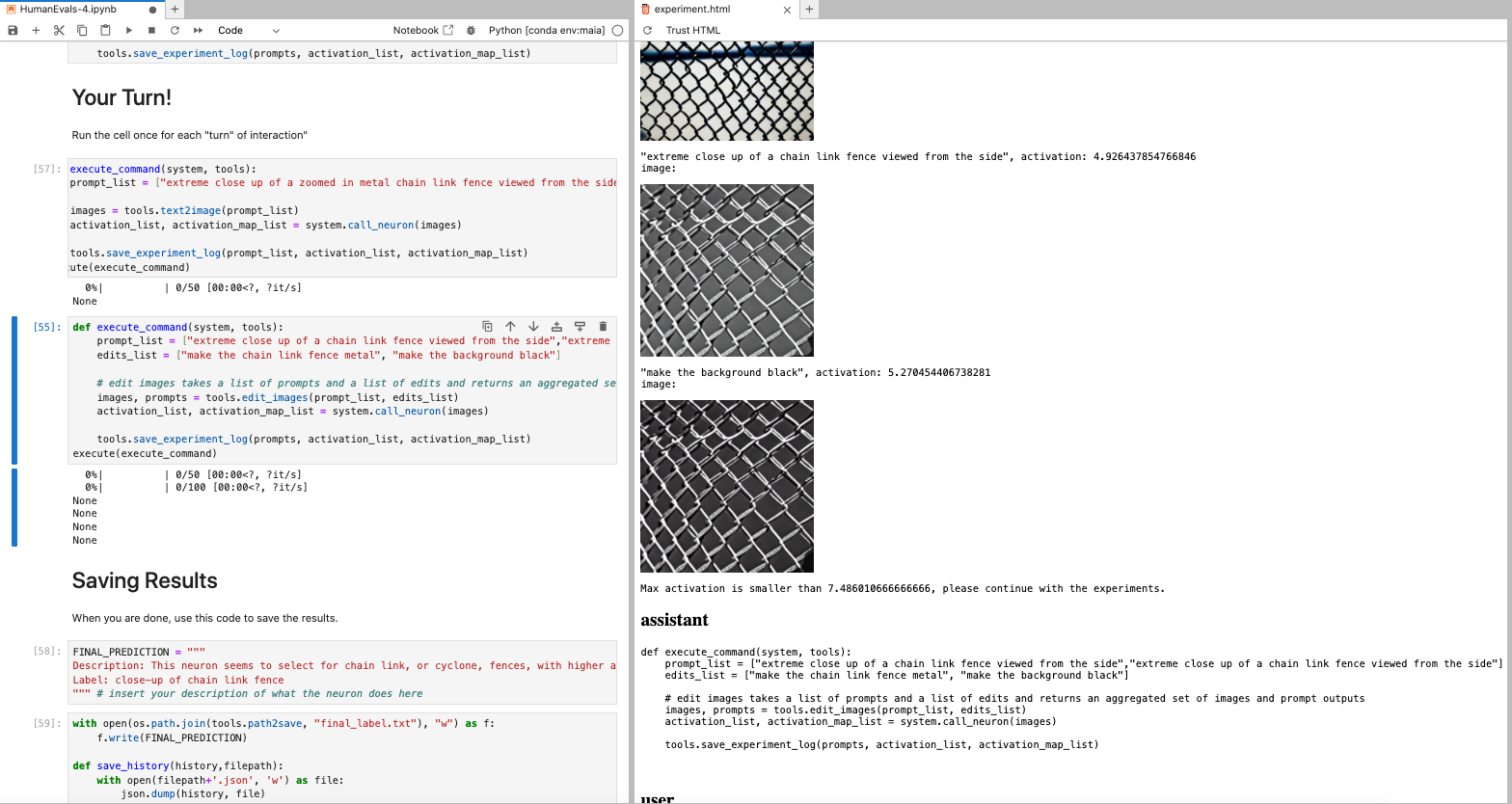}
  \caption{Example interface for humans interpreting neurons with the same tool library used by \maia.}
  \label{fig:human_notebook}
\end{wrapfigure}We recruited 8 human interpretability researchers to use the \maia API to run experiments on neurons in order to describe their behavior. 
 This data collection effort was approved by MIT's Committee on the Use of Humans as Experimental Subjects. Humans received task specification via the \maia user prompt, wrote programs using the functions inside the \maia API, and produced neuron descriptions in the same format as \maia. All human subjects had knowledge of Python. Humans labeled 25\% of the units in each layer labeled by \maia (one human label per neuron). Testing was administered via JupyterLab \citep{Kluyver2016jupyter}, as displayed in Figure \ref{fig:human_notebook}. Humans also labeled 25\% of the synthetic neurons using the same workflow. The median number of interactions per neuron for humans was 7. However, for more difficult neurons the number of interactions were as high as $39$.

\subsection{Synthetic neurons}
\label{sec:syn_appendix}

To provide a ground truth against which to test MAIA, we constructed a set of synthetic neurons that reflect the diverse response profiles of neurons in the wild. We used three categories of synthetic neurons with varying levels of complexity: \textit{monosemantic} neurons that respond to single concepts, \textit{polysemantic} neurons that respond to logical disjunctions of concepts, and \textit{conditional} neurons that respond to one concept conditional on the presence of another. The full set of synthetic neurons across all categories is described in Table \ref{tab:synthetic-list}. To capture real-world neuron behaviors, concepts are drawn from \milannotations, a dataset of 60K human annotations of prototypical neuron behaviors \citep{hernandez2022natural}.

\begin{table}[ht]
\centering
\caption{\textbf{Synthetic neurons.} Concepts are drawn from \milannotations.}
\label{tab:synthetic-list}
{\small %
\begin{tabular}{p{0.15\linewidth}p{0.25\linewidth}p{0.20\linewidth}}
\toprule
\textit{Monosemantic} & \textit{Polysemantic} (\textit{A} \verb|OR| \textit{B}) & \textit{Conditional} (\textit{A$|_{\text{B}}$}) \\
\midrule
arch & animal, door & ball, hand \\
bird & animal, ship & beach, people \\
blue & baby, dog & bird, tree \\
boat & bird, dog & bridge, sky \\
brick & blue, yellow & building, sky \\
bridge & bookshelf, building & cup, handle \\
bug & cup, road & dog, leash \\
building & dog, car & fence, animal \\
button & dog, horse & fish, water \\
car window & dog, instrument & grass, dog \\
circle & fire, fur & horse, grass \\
dog & firework, whisker & instrument, hand \\
eyes & hand, ear & skyline, water \\
feathers & necklace, flower & sky, bird \\
flame & people, building & snow, road \\
frog & people, wood & suit, tie \\
grass & red, purple & tent, mountain \\
hair & shoe, boat & water, blue \\
hands & sink, pool & wheel, racecar \\
handle & skirt, water & \\
hat & stairs, fruit & \\
jeans & temple, playground & \\
jigsaw & truck, train & \\
legs & window, wheel & \\
light &  & \\
neck &  & \\
orange &  & \\
paws &  & \\
pencil &  & \\
pizza &  & \\
roof &  & \\
shirt &  & \\
shoes &  & \\
sky &  & \\
snake &  & \\
spiral &  & \\
stripes &  & \\
sunglasses &  & \\
tail &  & \\
text &  & \\
tires &  & \\
tractor &  & \\
vehicle &  & \\
wing &  & \\
yarn &  & \\
\bottomrule
\vspace{-8mm}
\end{tabular}
}%
\end{table}

Synthetic neurons are constructed using Grounded DINO \citep{liu2023grounding} in combination with SAM \citep{kirillov2023segany}. Specifically, Grounded-DINO implements open-vocabulary object detection by generating image bounding boxes corresponding to an input text prompt. These bounding boxes are then fed into SAM as a soft prompt, indicating which part of the image to segment. To ensure the textual relevance of the bounding box, we set a threshold to filter out bounding boxes that do not correspond to the input prompt, using similarity scores which are also returned as synthetic neuron activation values. We use the default thresholds of 0.3 for bounding box accuracy and 0.25 for text similarity matching, as recommended in \citep{liu2023grounding}. After the final segmentation maps are generated, per-object masks are combined and dilated to resemble outputs of neurons inside trained vision models, instrumented via \maia's \code{System} class.

We also implement compound synthetic neurons that mimic polysemantic neurons found in the wild (via logical disjunction), and neurons that respond to complex combinations of concepts (via logical conjunction). To implement polysemantic neurons (\eg selective for \textit{A} \verb|OR| \textit{B}), we check if at least one concept is present in the input image (if both are present, we merge segmentation maps across concepts and return the mean of the two activation values). To implement conditional neurons (\eg selective for \textit{A}$|_{\text{B}}$), we check if $A$ is present, and if the condition is met ($B$ is present) we return the mask and activation value corresponding to concept \textit{A}.

The set of concepts that can be represented by synthetic neurons is limited by the fidelity of open-set concept detection using current text-guided segmentation methods. We manually verify that all concepts in the synthetic neuron dataset can be consistently segmented by Grounded DINO in combination with SAM. There are some types of neuron behavior, however, that cannot be captured using current text-guided segmentation methods. Some neurons inside trained vision models implement low-level procedures (\eg edge-detection), or higher level perceptual similarity detection (\eg sensitivity to radial wheel-and-spoke patterns common to flower petals and bicycle tires) that Grounded DINO + SAM cannot detect. Future implementations could explore whether an end-to-end single model open-vocabulary segmentation system, such as Segment Everything Everywhere All at Once \citep{zou2023segment}, could perform segmentation for richer neuron labels.

\paragraph{Evaluation of synthetic neuron labels using human judges.}
This data collection effort was approved by MIT’s Committee on the Use of Humans as Experimental Subjects. To control for quality, workers were required to have a HIT acceptance rate of at least 99\%, be based in the U.S., and have at least 10,000 approved HITs. Workers were paid $0.10$ per annotation. 10 human judges performed each comparison task. 

\newpage

\begin{wrapfigure}{r}{0.52\textwidth}
  \centering
  \vspace{-.5cm}
  \includegraphics[width=1\linewidth]{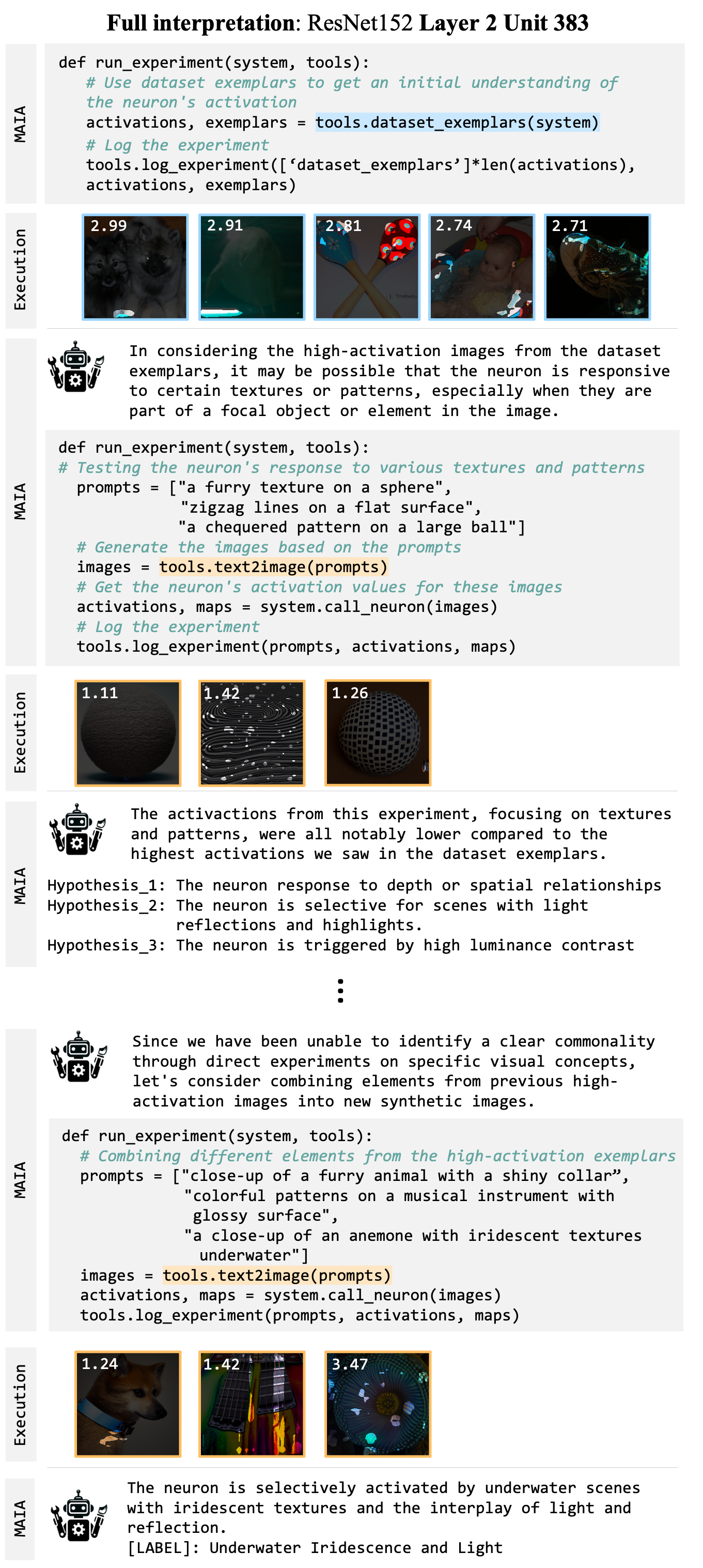}
  \caption{Example of \maia having confirmation bias towards a single generated example, instead of generating further experiments to test other possibilities.}
  \label{fig:overfitting}
  \vspace{-3.1cm}
\end{wrapfigure}


\vspace{5mm}
\section{Failure modes}
\label{sec:failures_appendix}

\subsection{Tool Failures}


\maia is often constrained by the capabilities of its tools. As shown in Figure \ref{fig:failures_main}, the Instruct-Pix2Pix \cite{brooks2022instructpix2pix} and Stable Diffusion \cite{rombach2022highresolution} models sometimes fail to follow the precise instructions in the caption. Instruct-Pix2Pix typically has trouble making changes which are relative to objects within the image and also fails to make changes that are unusual (such as the example of replacing a person with a vase). Stable Diffusion typically has difficulty assigning attributes in the caption to the correct parts of the image. These errors in image editing and generation sometimes confuse \maia and cause it to make the wrong prediction.

\subsection{Confirmation Bias}
In some scenarios, when \maia generates an image that has a higher activation than the dataset exemplars, it will assume that the neuron behaves according to that single exemplar. Instead of conducting additional experiments to see if there may be a more general label, \maia sometimes stops experimenting and outputs a final label that is specific to that one image. For instance, in Figure \ref{fig:overfitting} \maia generates one underwater image that attains a higher activation and outputs an overly specific description without doing any additional testing.

\subsection{Failure modes of non-proprietary VLMs as alternative MAIA backbones.}

We tested the performance of two models as the MAIA backbone VLM:
\begin{enumerate}
    \item LLaVA-Next~\cite{liu2024llavanext}, a top-performing (improved over LLaVA-1.5) open source VLM built from LLaMA. 
    \item Gemini 1.0 Pro~\cite{team2023gemini}, which Google currently provides for free with a low rate limit (60 RPM). 
\end{enumerate}

When used as the MAIA backbone for the neuron interpretation task, both models show initial potential: they understand the task, use the System class and the Tools class from MAIA’s API, and write code specifying interpretation experiments. However, while both models show promise, we found significant shortcomings compared to GPT-4V which limit their current usefulness as backbones of MAIA. Some of these shortcomings we observed in both models:
 
\textbf{Weaker hypothesis generation.} Key to MAIA is the ability of the backbone VLM to make and update hypotheses in light of experimental findings. After receiving image outputs of experiments, both LLaVA-Next and Gemini are biased toward describing the images rather than updating their hypothesis about system behavior (suggesting both were trained for the task of image captioning, which might be solvable with fine-tuning). 

\textbf{Hallucination.} We observed that both models often output hypotheses unrelated to experimental outcomes, and sometimes hallucinate experimental results.
For example, when Gemini got the activation value and masked image for the prompt \textit{``a dog standing on the grass''} it replied with: 
\texttt{a cat laying on the grass, activation: 14.21},
which is an image never generated, and a hallucinated activation value.

\textbf{Overfitting to the examples of the tool usage presented in the MAIA API.} Rather than designing new experiments, both models often reproduce code that appears in the MAIA API usage examples.  
For example, both Llava and Gemini will generate images of \textit{``a dog standing on the grass''} which is provided as an example for an input prompt to the text2image tool. 

\textbf{Gemini did show some advantages over LLaVA:} while Gemini occasionally produced some syntax errors (\eg calling \texttt{system.dataset\_exemplars} instead of \texttt{tools.dataset\_exemplars}), Llava code was often not executable. Furthermore, LLaVa is restricted to getting only one input image at each interaction, which severely restricts its experimentation ability. Even when inputting several images in a row, the model is biased toward analyzing the last one.

\label{app:vlms}

\section{Utility of \maia descriptions for human users}
\label{sec:mturk2}
We run additional crowd-sourcing experiments to quantify the extent to which neuron descriptions equip humans to predict system behavior. Given a text description of a neuron (produced either by \milan, \maia or human experts), human participants predicted the neuron’s activations on images from the ImageNet validation set. In this setting, a correct description would inform more accurate predictions of behavior. 

Specifically, participants saw a language description of a neuron (e.g. “this neuron is selective for human hands interacting with weightlifting equipment”) and four sets of images. One of those sets contains images that strongly activate the neuron described in text, and the other 3 contain randomly sampled distractors (capturing ``baseline activity''). Humans are asked to use the text description to select the strongly activating set of images (chance is 25\%). 

Two notes on experimental design: (i) We used sets of images (4 images each) in the multiple choice task instead of single images to provide better coverage of concept space (some text labels describe more concepts than a single image could show). (ii) In this task, humans distinguish strongly activating images from weakly activating images (like the automated evaluation in the main paper), instead of ordering images along a continuum by how strongly they activate the neuron. This is because the explanations produced by the interpretability methods only describe concepts that strongly activate the neuron, and do not provide enough information for human observers to characterize the full distribution of the neuron’s activity. 

The table below shows results from this experiment and 95\% confidence intervals across interpretation procedures for a given model. For each model we evaluated the same subset of 25 neurons as in Section ~\ref{sec:evaluation}. 10 human participants performed each task (using the same crowdworker selection criteria as in Section ~\ref{sec:syn_eval}). \maia descriptions are more useful for predicting neuron behavior than \milan descriptions in all three models studied.

\begin{table}[h]
\centering
\vspace{-4mm}
\caption{Subject predictions of maximally activating images based on neuron descriptions.}
\begin{tabular}{lccc}
\toprule
 & \milan & \maia & Human \\ \hline
ResNet-152 & 66.0 $\pm$ 0.06 & 78.5 $\pm$ 0.05 & 85.45 $\pm$ 0.04 \\
CLIP-RN50 & 57.19 $\pm$ 0.06 & 65.21 $\pm$ 0.06 & 69.6 $\pm$ 0.06 \\ 
DINO-ViT & 48.0 $\pm$ 0.06 & 67.39 $\pm$ 0.06 & 81.2 $\pm$ 0.05 \\ 
\bottomrule
\end{tabular}
\label{tab:comparison}
\end{table}

\section{Spurious feature removal experiment}
\label{sec:spurious_appendix}

\subsection{Dataset Selection}
We use the Spawrious dataset as it provides a more complex classification task than simpler binary spurious classification benchmarks like Waterbirds \cite{wah_branson_welinder_perona_belongie_2011, sagawa2020distributionally} and CelebA \cite{liu2015deep, sagawa2020distributionally}. All images in the dataset are generated with Stable Diffusion v1.4 \cite{rombach2022highresolution}, which is distinct from the Stable Diffusion v1.5 model in \maia's tool API. See \citet{lynch2023spawrious} for further specific details on dataset construction.

\begin{figure}[b]
  \centering
  \begin{minipage}{.46\textwidth}
    \centering
    \includegraphics[width=.8\linewidth]{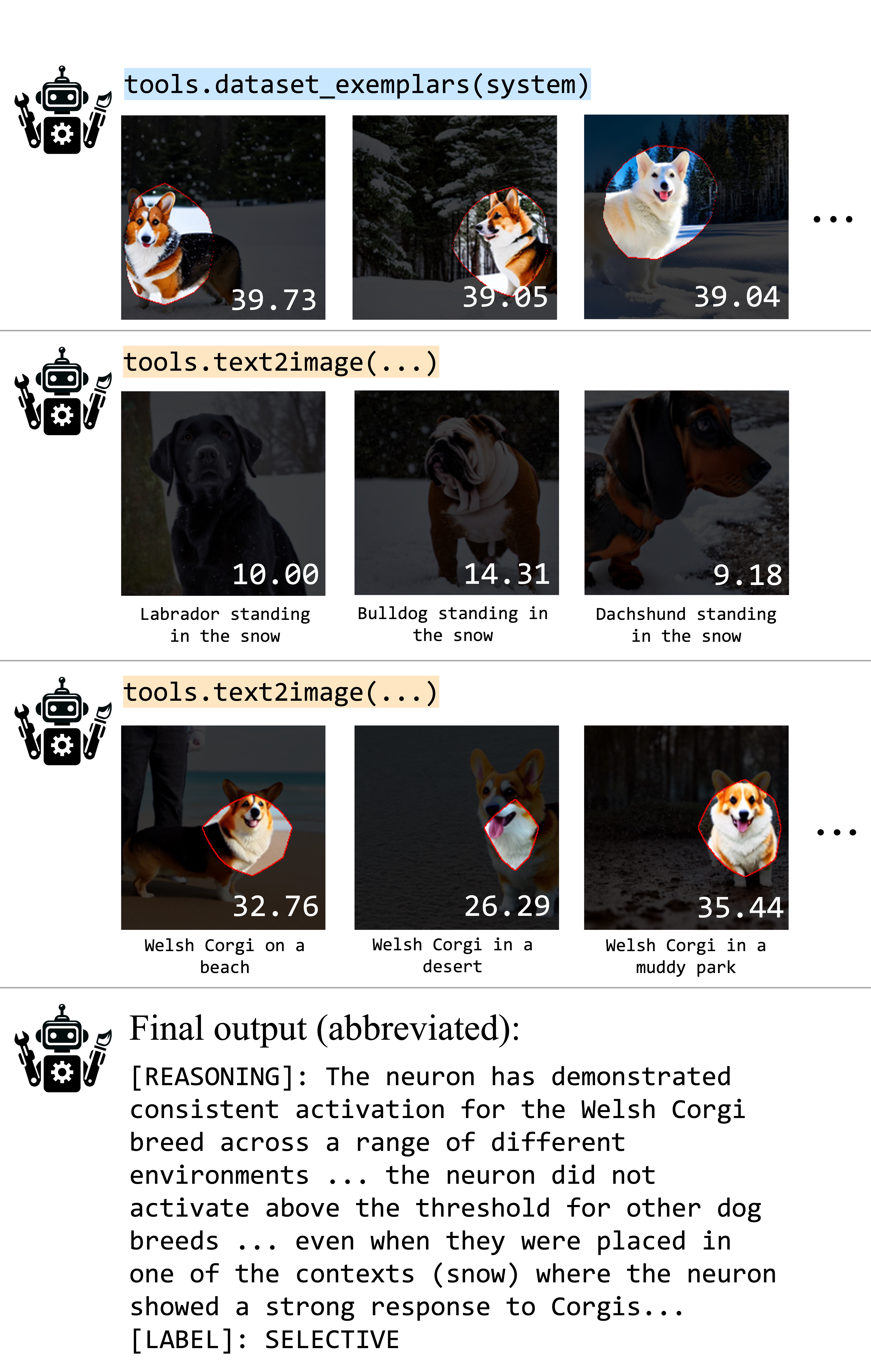}
  \end{minipage}
  \hspace{0.02\textwidth}
  \begin{minipage}{.46\textwidth}
    \centering
    \includegraphics[width=.95\linewidth]{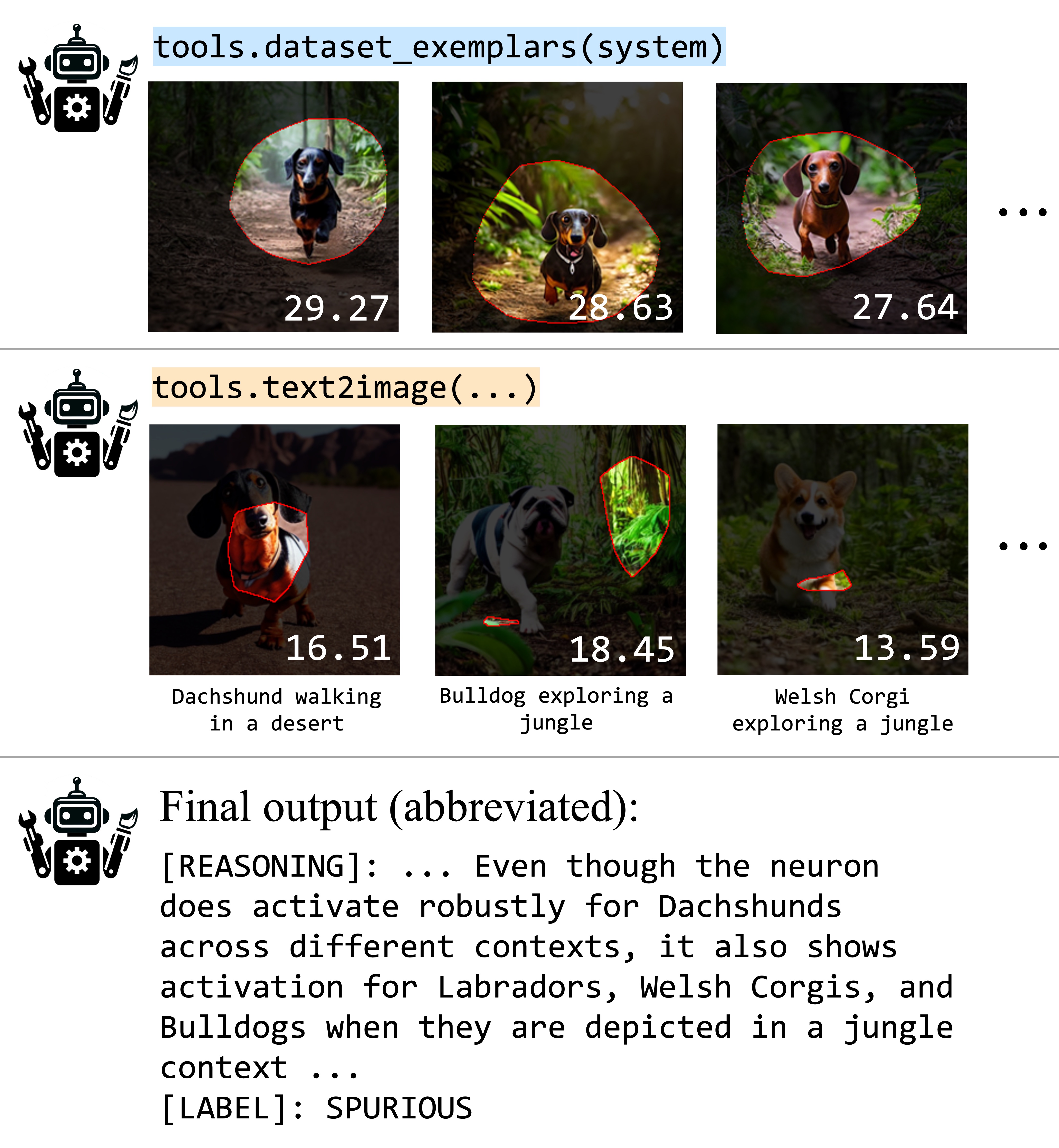}
  \end{minipage}
 \centering
 \begin{minipage}{0.9\textwidth}
   \centering
   \caption{Two different \maia interactions, classifying neurons as selective (\emph{left}) and spurious (\emph{right}).}
   \label{fig:good_selective_spurious}
 \end{minipage}
\end{figure}

\subsection{Experiment Details} \label{sec:spaw_experiment_details}
Here, we describe the experiment details for each row from Table \ref{tab:spawrious}. Note that for all the logistic regression models that we train, we standardize the input features to have a zero mean and variance of one. We use the \verb|saga| solver from \verb|sklearn.linear_model.LogisticRegression| for the $\ell_1$ regressions and the \verb|lbfgs| solver for the unregularized regressions \cite{scikit-learn}.

\textbf{All, Original Model, Unbalanced:} We train a ResNet-18 model \cite{he2016deep} for one epoch on the O2O-Easy dataset from Spawrious using a learning rate of 1e-4, a weight decay of 1e-4, and a dropout of 0.1 on the final layer. We use a 90-10 split to get a training set of size 22810 and a validation set of size 2534.

\textbf{$\ell_1$ Top 50, All, Unbalanced:} We tune the $\ell_1$ regularization parameter on the full unbalanced validation set such that there are $50$ neurons with non-zero weigths, and we extract the corresponding neurons indices. We then directly evaluate the performance of the logistic regression model on the test set.

\textbf{$\ell_1$ Top 50, Random, Unbalanced:} To match \maia's sparsity level, we extract 100 sets of 22 random neuron indices and perform unregularized logistic regression on the unbalanced validation set.

\textbf{$\ell_1$ Top 50, $\ell_1$ Top 22, Unbalanced:} We also use $\ell_1$ regression to match \maia's sparsity in a more principled manner, tuning the $\ell_1$ parameter until there are only $22$ neurons with non-zero weights. We then directly evaluate the performance of the regularized logistic regression model on the test set.

\textbf{$\ell_1$ Top 50, \milan, Unbalanced:} We use \milan to caption all 50 neurons. Since \milan was not trained to annotate neurons as selective or spurious, we manually select all neurons with captions related to dogs (and not to their backgrounds). After filtering, this set contains 23 neurons. We then perform unregularized logistic regression with this neuron subset on the unbalanced validation set.

\textbf{$\ell_1$ Top 50, \milan (GPT-4V), Unbalanced:} We repeat the same process as in \textbf{$\ell_1$ Top 50, \milan, Unbalanced}, but this time we use GPT-4V to annotate maximally activating exemplars instead of the original \milan model. This subset contains 23 neurons. We then perform unregularized logistic regression with this neuron subset on the unbalanced validation set. 

\textbf{$\ell_1$ Top 50, \maia, Unbalanced:} We run $\maia$ on each of the 50 neurons separately, and it ultimately selects 22 out of the 50 neurons. We then perform unregularized logistic regression with this neuron subset on the unbalanced validation set. We use a modified user prompt which we include in Section \ref{sec:spaw_prompting}.

Next, for the balanced validation fine-tuning experiments, we sample ten balanced validation sets of size 320 and report the mean performances of each method. While \citet{kirichenko2023layer} uses multiple subsampled balanced validation sets for fine-tuning and then aggregates the models for scoring on the test set, we only allow the following experiments to see a \textit{single} balanced validation set since we seek to compare the performance of \maia to methods which have access to a small balanced fine-tuning dataset, rather than spurious feature labels for every data sample in a large validation set.

\textbf{All, $\ell_1$ Tuning, Balanced:} We perform a hyperparameter search for the $\ell_1$ parameter, evaluating each hyperparameter value with five 50-50 splits of the balanced validation dataset, training on one half and evaluating on the other half. We then train the final model with the best $\ell_1$ parameter on the entire balanced validation dataset. For the number of neurons, we report the median number of neurons with non-zero weights across the ten trials.

\textbf{All, $\ell_1$ Top 22, Balanced:} We also investigate the performance when we match \maia's chosen neuron sparsity level by tuning the $\ell_1$ parameter to only have 22 neurons with non-zero weights. We train the model directly on the balanced validation dataset.

\subsection{Example \maia Neuron Robustness Interactions} \label{sec:spaw_examples}

In Figure \ref{fig:good_selective_spurious}, we show examples of \maia's interaction with neurons in the final layer of the Resnet-18 model trained on Spawrious. Overall, we find that \maia will often test all four dog breeds across different environments, and if it finds that the exemplars all have the same background, it will test to see if the background is a spurious feature.


\subsection{\maia Prompting Details}\label{sec:spaw_prompting}

The user prompt to \maia specifying the spurious neuron identification task is similar to the prompt specifying the neuron description task. It is shown in full below. We only specify to \maia the domain of images the model was trained on and the definitions of spurious and selective neurons so that it can output a final classification using experiments that will not generate significantly out-of-distribution images (such as unrelated dog breeds or close-up poses not present in the training data). We do not ever specify what the spurious correlations are.

\begin{mygraybox}{}
You are analyzing the prototypical behavior of individual neurons inside a deep neural network which classifies the breed of a dog in a natural image as one of the following breeds that were in its training dataset:
1. Labrador
2. Welsh Corgi
3. Bulldog
4. Dachshund
Your overall task is to classify the neuron as SELECTIVE (if it is selective for one and only one dog breed) or SPURIOUS (if it is not). Conduct experiments until you meet the following criteria for SELECTIVE or SPURIOUS. 

SELECTIVE
Criteria:
1. The neuron only activates for one single breed of dog. It does not activate for any other breeds of dog, even at lower magnitudes of activation. We consider a neuron to be activated if it returns a highlighted image region with a red outline. Observe the presence or absence of a mask, and the magnitude of an activation value, to determine whether a neuron is active. 
2. The neuron is active for that single dog breed robustly across a variety of contexts. However, keep in mind that the neuron has only seen images of full-body dogs in the following environments: beach, desert, muddy park, jungle, and snow. Do not generate images that are very dissimilar to what the neural network has seen before (eg. do not generate close-ups, do not generate humans). A selective neuron does not need to activate for these dissimilar images.

SPURIOUS
Criteria:
1. The neuron activates for mutiple breeds of dogs. The amount the neuron activates for each breed of dog may vary but the neuron is considered spurious if it activates enough for multiple breeds to return a highlighted image region with a red outline.
2. The neuron may activate for a single breed of dog, but only does so in specific contexts. If the neuron's activation is dependent on non-dog related concepts, it is considered spurious. 

To complete your task, you are provided with a library of Python functions to run experiments on the specific neuron (inside the "System" class) given the functions provided in the "Tools" class. Make sure to use a variety of tools from the library to maximize your experimentation power.
Some neurons might be selective for very specific concepts, a group of unrelated concepts, or a general concept, so try to be creative in your experiment and try to test both general and specific concepts. If a neuron is selective for multiple concepts, you should describe each of those concepts in your final description. At each experiment step, write Python code that will conduct your experiment on the tested neuron, using the following format:
[CODE]: 
```python
def execute_command(system, tools)
    # gets an object of the system class, an object of the tool class, and performs experiments on the neuron with the tools
    ...
    tools.save_experiment_log(...)
```
Finish each experiment by documenting it by calling the "save_experiment_log" function. Do not include any additional implementation other than this function. Do not call "execute_command" after defining it. Include only a single instance of experiment implementation at each step. Each time you get the output of the neuron, try to summarize what inputs that activate the neuron have in common (where that description is not influenced by previous hypotheses), and make a hypothesis regarding whether the neuron is SELECTIVE (activates strongly for only one dog breed across contexts) or SPURIOUS (activates for more than one dog breed or non-dog related features). After each experiment, wait to observe the outputs of the neuron. Then your goal is to draw conclusions from the data, update your hypothesis, and write additional experiments to test it. Test the effects of concepts like different dog breeds or background features using the different tools in the library. If you are unsure about the results of the previous experiment you can also rerun it, or rerun a modified version of it with additional tools.
Use the following format:
[HYPOTHESIS]: ## update your hypothesis according to the image content and related activation values. Only update your hypotheses if image activation values are higher than previous experiments.
[CODE]: ##Test different and specific aspects of your hypothesis using all of the tools in the library. Write code to run the experiment in the same format provided above. Include only a single instance of experiment implementation.

Continue running experiments until you meet one of the following stopping criteria.
SPURIOUS: If you find multiple pieces of evidence which show that the neuron does not only activate for one breed or activates for non-dog related concepts you should deem the neuron to be spurious, even if you are not entirely sure what the neuron activates for. Remember that we consider a neuron to have activation if it returns a highlighted image region with a red outline. If you see this happen for a feature that is not dog related or for multiple breeds you should deem the neuron to be spurious.
SELECTIVE: The neuron is selective for a single dog breed and activates strongly for that breed across contexts. If you find any evidence suggesting that a neuron is spurious (such as an image outside one particular breed showing activations), you should conduct more experiments to test your hypotheses. If you are ever unsure about the result, you should lean towards outputting SPURIOUS. The neuron must be perfectly selective across many variations of dog breeds and contexts to be considered to be selective. Once you have met one of the stopping criteria, output your final classification of the neuron in the following format:
[REASONING]: <why the neuron is SELECTIVE or SPURIOUS, and if SELECTIVE, the breed it is SELECTIVE for>
[LABEL]: <SELECTIVE/SPURIOUS>
\end{mygraybox}

\section{Bias identification experiment}
\label{sec:bias-appendix}
The user prompt to \maia specifying the bias identification task is similar to the prompt specifying the neuron description task. We produce it in full below. 

\begin{mygraybox}{}
Your overall task is to describe the visual concepts that maximally activate a neuron inside a deep network for computer vision. 
You will get an access to a neuron inside a deep neural network that is sensitive to the presence of {class_label} inside the input image. Your overall task is to reveal biases in the neuron's behavior. That is, your task is to find specific contexts or settings in which the neuron tends to have relatively higher activation values, but only covers partial asspect of the {class_label} class, or specific scenarios when the neuron tends to have relatively lower activations although {class_label} is still present. 
Try to look for different type of biases (e.g. gender, ethnicity, context-dependencies, breed-dependencies, etc.). 
To do that you are provided with a library of Python functions to run experiments on the specific neuron (inside the "System" class) given the functions provided in the "Tools" class. Make sure to use a variety of tools from the library to maximize your experimentation power. Some neurons might be selective for very specific concepts, a group of unrelated concepts, or a general concept, so try to be creative in your experiment and try to test both general and specific concepts. If a neuron is selective for multiple concepts, you should describe each of those concepts in your final description. At each experiment step, write Python code that will conduct your experiment on the neuron, using the following format:
[CODE]: 
```python
def execute_command(system, tools)
    # gets an object of the system class, an object of the tool class, and performs experiments on the neuron with the tools
    ...
    tools.save_experiment_log(...)
```
Finish each experiment by documenting it by calling the "save_experiment_log" function. Do not include any additional implementation other than this function. Do not call "execute_command" after defining it. Include only a single instance of experiment implementation at each step.

Each time you get the output of the neuron, try to summarize what inputs that activate the neuron have in common (where that description is not influenced by previous hypotheses). Then, write multiple hypotheses that could explain the biases of the neuron.
For example, these hypotheses could list multiple context that the neuron is less selective for.
Then write a list of initial hypotheses about the neuron biases in the format:
[HYPOTHESIS LIST]: 
Hypothesis_1: <hypothesis_1>
...
Hypothesis_n: <hypothesis_n>.

After each experiment, wait to observe the outputs of the neuron. Then your goal is to draw conclusions from the data, update your list of hypotheses, and write additional experiments to test them. Test the effects of both local and global differences in images using the different tools in the library. If you are unsure about the results of the previous experiment you can also rerun it, or rerun a modified version of it with additional tools.
Use the following format:
[HYPOTHESIS LIST]: ## update your hypothesis list according to the image content and related activation values. 
Only update your hypotheses if image activation values are higher than previous experiments.
[CODE]: ## conduct additional experiments using the provided python library to test *ALL* the hypotheses. Test 
different and specific aspects of each hypothesis using all of the tools in the library. Write code to run 
the experiment in the same format provided above. Include only a single instance of experiment implementation.

Continue running experiments until you prove or disprove all of your hypotheses. Only when you are confident in your hypothesis after proving it in multiple experiments, output your final description of the neuron in the following format:

[BIAS]: <final description of the neuron bias>

\end{mygraybox}
}


\end{document}